%% file: main.tex
\documentclass[10pt,twocolumn,letterpaper]{article}

\usepackage[pagenumbers]{cvpr} 
\usepackage{float}
\input{preamble}

\definecolor{cvprblue}{rgb}{0.21,0.49,0.74}
\usepackage[pagebackref,breaklinks,colorlinks,citecolor=cvprblue]{hyperref}

\def\paperID{1312} 
\def\confName{CVPR}
\def\confYear{2024}
\usepackage{multirow}

\title{Navigating Beyond Dropout: An Intriguing Solution Towards Generalizable Image Super Resolution
}

\author{Hongjun Wang\thanks{Equal Contribution} \\
	The University of Tokyo\\
	{\tt\small hjwang-ai@g.ecc.u-tokyo.ac.jp}
\and
Jiyuan Chen\footnotemark[1]\\
The Hong Kong Polytechnic University\\
{\tt\small jiyuan.chen@connect.polyu.hk}
\and 
Yinqiang Zheng\thanks{Corresponding Author}\\
The University of Tokyo\\
{\tt\small yqzheng@ai.u-tokyo.ac.jp}
\and
Tieyong Zeng\footnotemark[2]\\
The Chinese University of Hong Kong\\
{\tt\small zeng@math.cuhk.edu.hk}
\and 
\textcolor{magenta}{\url{https://github.com/Dreamzz5/Simple-Align}} 
\and 
}

\newtheorem{lemma}{Lemma}[section]

\newcommand{\secref}[1]{Sec.~\ref{#1}}
\newcommand{\tableref}[1]{Table~\ref{#1}} 

\newcommand{\figref}[1]{Fig.~\ref{#1}} 
\usepackage{bbm}
\begin{document}
\maketitle
\begin{abstract}
	
	
	Deep learning has led to a dramatic leap on Single Image Super-Resolution (SISR) performances in recent years. 
	While most existing work assumes a simple and fixed degradation model (e.g., bicubic downsampling), the research of Blind SR seeks to improve model generalization ability with unknown degradation. Recently, \citet{kong2022reflash} pioneer the investigation of a more suitable training strategy for Blind SR using Dropout \cite{tompson2015efficient}. Although such method indeed brings substantial generalization improvements via mitigating overfitting, we argue that Dropout simultaneously introduces undesirable side-effect that compromises model's capacity to faithfully reconstruct fine details. We show both the theoretical and experimental analyses in our paper, and furthermore, we present another easy yet effective training strategy that enhances the generalization ability of the model by simply modulating its first and second-order features statistics. Experimental results have shown that our method could serve as a model-agnostic regularization and outperforms Dropout on seven benchmark datasets including both synthetic and real-world scenarios.


\end{abstract}
\section{Introduction}
\label{sec:intro}

\begin{figure}
	\centering
	\includegraphics[width=1\linewidth]{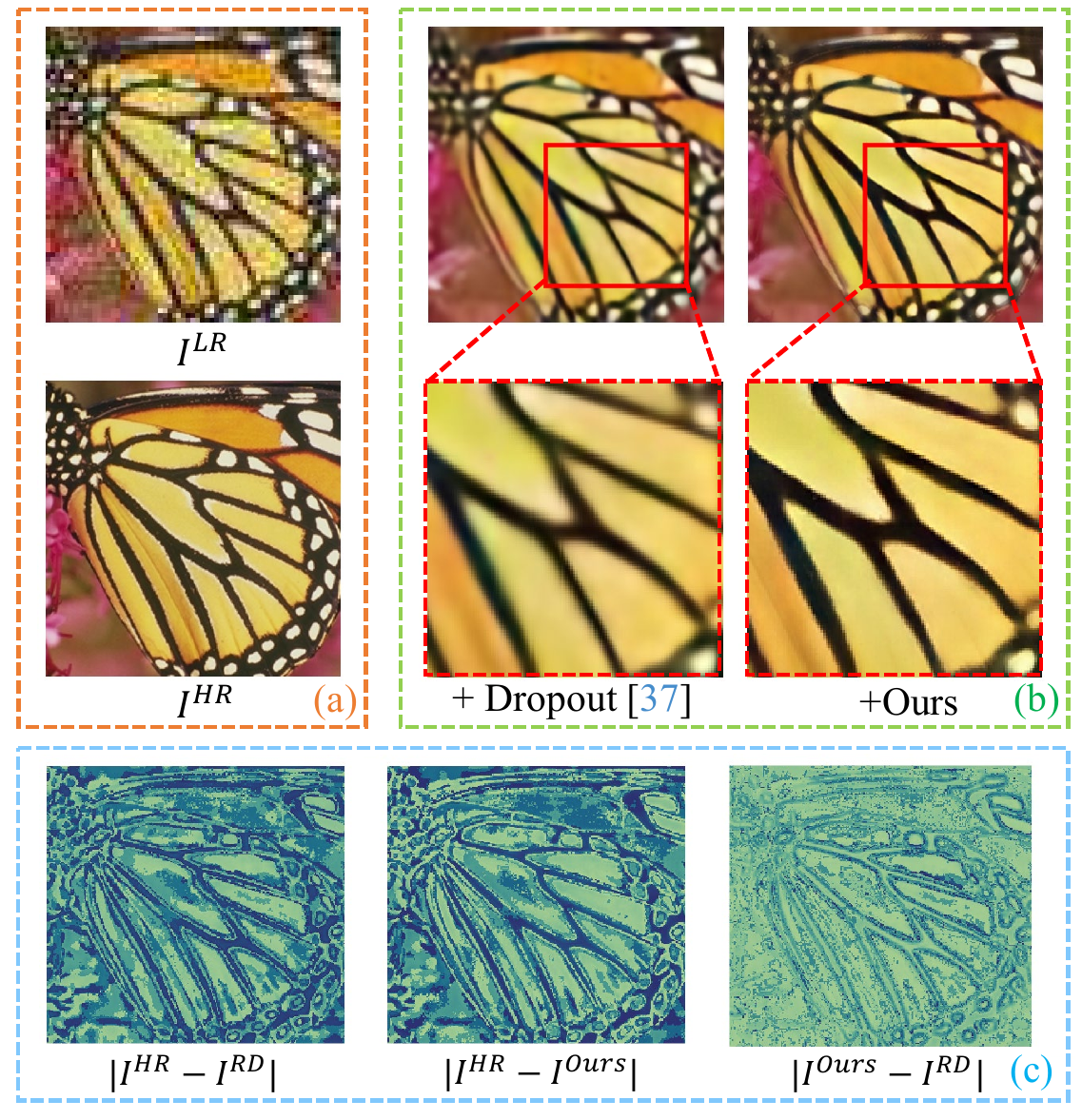}
	\caption{Given a HR image $I^{HR}$ and its LR version $I^{LR}$ in part (a), the visual comparisons of the restored results of SRResNet \cite{SRResNet}, regularized by Dropout \cite{kong2022reflash} and our method respectively, are shown in part (b). Part (c) presents the estimated residuals. \textbf{We could observe that our method gives better visual quality and preserves more vivid details.}}
	\label{fig:motivation}
\end{figure}

Riding on the waves of the explosive advancement of deep learning technology, Single Image Super-Resolution (SISR) with deep neural networks (DNNs) has greatly evolved in recent years (e.g., VDSR \cite{VDSR}, SRResNet \cite{SRResNet}, EDSR \cite{EDSR}, RDN \cite{RDN} and SwinIR \cite{liang2021swinir}), offering superior performances over traditional prediction models \cite{irani1991improving,fattal2007image,shan2008fast,glasner2009super}.


However, due to the non-trivial issue  of collecting a massive amount of natural low-resolution (LR) and high-resolution (HR) image pairs for DNN's training \cite{wang2021real,liu2022blind}, early researchers of SISR resort to manually designed HR/LR image pairs (i.e., bicubic) as a surrogate.  Nevertheless, realistic degradations barely obey such simple assumption, leading to severe performance drop for these models.


Blind SR \cite{liu2022blind,wang2021unsupervised,zhang2021designing,ji2020real,yuan2018unsupervised}, as an answer to the above question, seeks to improve model generalization ability with unknown degradations. 
Despite promising results have been obtained with enriching training degradation space (e.g., through handcrafted synthesisation \cite{zhang2021designing, wang2021real, sahak2023denoising} or data distribution learning \cite{chen2023better,li2022face, bulat2018learn}), and enhancing model capability (e.g., unfolding degradation model \cite{huang2020unfolding,zheng2022unfolded} or exploring image internal statistics \cite{shocher2018zero}), we notice that the investigation of training strategy (regularization) 
that benefits Blind SR has been barely touched so far.

In this paper, we argue that such investigation is necessary and meaningful, given current research status of Blind SR. The reasons are as follows. Firstly, without enlarging degradation space for training, the development of Blind SR has reached its bottleneck \cite{liu2022blind,li2022face}. Even though there are methods trying to excavate image's internal similar patterns to perform zero-shot learning \cite{cheng2020zero,shocher2018zero}, they can easily fail in natural cases where self-repeated patterns are absent. On the other hand, training with a large degradation space theoretically grants the model better generalization ability
by encouraging it to focus more on learning the shape and texture prior of natural images. Therefore, constructing a diverse degradation pool for training (namely, data-driven based Blind SR, according to \citet{li2022face}) has become a mainstream direction for recent Blind SR researches \cite{chen2023human,chen2023better,bulat2018learn,wei2021unsupervised,bell2019blind,ji2020real,zhou2019kernel}, and has been shown effective in both CNN-based \cite{zhang2018learning,liang2022efficient}, GAN-based \cite{zhang2021designing,wang2021realesrgan} and Diffusion-based \cite{yang2023synthesizing,sahak2023denoising} models. 

However, only under ideal circumstances that the model trained with diverse degradations can automatically unleash its full potential to learn degradation invariant representation \cite{li2023learning}, and thus becomes more generalizable to unknown degradations. Experiences from other fields \cite{gu2022clothes,qin2021causal,cadene2019rubi,zhao2022learning} warn that without proper regularization, such ideal case might not be so easily achieved.
Recently, \citet{kong2022reflash} first noticed this problem and refreshed the usage of Dropout to mitigate the ``overfitting to degradation" issue in data-driven Blind SR. They pointed out that without proper regularization, simply increasing the data and network scale can not continuously improve generalization ability now. Nevertheless, we notice that despite the enhanced performances, Dropout also introduces unwanted side-effects that reduce feature interaction and diversity, which further leads to the loss of vivid high frequency details. We show a preliminary example in \figref{fig:motivation}, and we will elaborate the theoretical and experimental analyses in \secref{sec:dropout}.


Moreover, the ``overfitting to degradation" problem stems from the network's excessive attention on some specific degradations. To tackle this issue, we further propose a statistical alignment method that during training, aligns the first and second order feature statistics (i.e., mean and covariance) of two images which have the same content but different degradations. We observe that such simple regularization can effectively enhance the model's ability to selectively remove degradation-related information during forward pass. Therefore, the model can recover fine-grained HR contents free from distraction of degradations and becomes more generalizable. Our regularization can easily cope with existing popular DNNs in a model-agnostic way and its implementation pre-request (i.e., images with the same content but different degradations) cooperates smoothly with current data-driven Blind SR methods which have stochastic degradation-generation models \cite{luo2022learning,zhou2019kernel,wang2021real,bell2019blind}. The details of our method will be introduced in \secref{sec:method} and we here summarize the main contributions of this paper as follows:

\begin{itemize}[leftmargin=*]
	\item[$\bullet$] We argue and verify both theoretically and experimentally that Dropout is  not a desirable regularization choice for Blind SR setting due to its side-effect in reducing feature interaction and diversity, which further leads to loosing information especially in high frequency details.
	
	
	\item[$\bullet$] We propose a simple statistical alignment method that encourage the model to be thoroughly unaware of degradation information, therefore excavating the full potential of model generalization ability. 
	Note that our regularization actually works in parallel and serves as a complement to existing data-driven Blind SR researches. 
	
	\item[$\bullet$] We conduct extensive experiments on seven widely used benchmarks to validate our proposals and arguments.
	
\end{itemize}

\section{Related Work}
\label{sec:related_work}




Blind image super-resolution aims to effectively restore HR images from their LR counterparts with unknown degradations. Over the years, the solutions of this question can be roughly categorized into three brunches. 
The first brunch of researchers seek to collect real-world HR-LR image pairs for training \cite{cai2019toward,zhang2019zoom,chen2019camera} by adapting the focal length of cameras. However, such collection process is cumbersome and prone to spatial misalignment, making the hope of building a large and diverse training set nearly impossible. 

Considering the difficulties faced by the abovementioned work, the second brunch of researchers completely remove the need for external data by performing zero-shot learning. Representative work in this direction includes ZSSR \cite{shocher2018zero} and DGDML-SR \cite{cheng2020zero}, which utilize bicubic downsampling and depth information as super-resolution priors respectively. However, these methods rely heavily on the frequently recurring contents of the image, constraining their favourable performances to a very limited set of data.

The core of the third brunch of work lies in enriching the training degradation space, either with hand-crafted synthesisation \cite{zhang2021designing, wang2021real, sahak2023denoising} or data distribution learning \cite{chen2023better,li2022face, bulat2018learn}. Such idea dates back to the very beginning of Blind SR research (i.e., SRMD \cite{zhang2018learning}), and is much in line with the instincts of machine learning that large training space naturally leads to better generalization. \citet{zhang2021designing} and \citet{wang2021real} proposed to use repeated synthesized degradations instead of single ones to build more generic datasets. Later on, to further stimulate real-world degradations, GAN \cite{bell2019blind,chen2023better,li2022face, bulat2018learn,lugmayr2019unsupervised,zhou2019kernel} and Diffusion models \cite{yang2023synthesizing} are incorporated to learn more realistic distributions.  For the stochasticity of degradation learning, \citet{bulat2018learn} and \citet{maeda2020unpaired} propose to integrate random vectors into degradation modeling, and \citet{luo2022learning} further design a unified probabilistic framework that has general applicability. We also categorize those work that explores model designs on the top of such multi-degradation settings into this brunch, including degradation-adaptive networks \cite{IKC,VDSR,liang2022efficient,wang2021unsupervised} and deep unfolding networks \cite{huang2020unfolding,zheng2022unfolded,zhang2020deep}.

\begin{figure}
	\centering
	\includegraphics[width=0.85\linewidth]{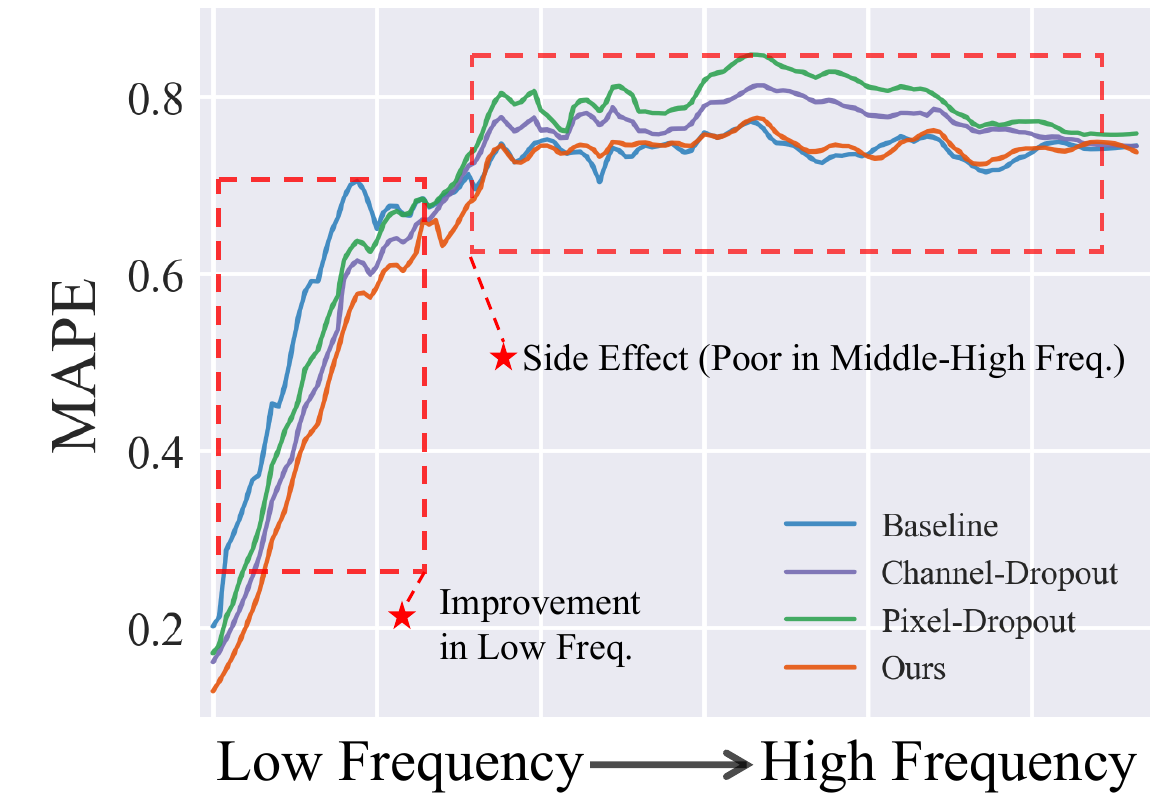}
	\vspace{-5pt}
	\caption{\textbf{The MAPE of SRResNet in frequency domain.} We can observe that both channel and pixel dropout have inferior performances to the pure SRResNet in middle-high frequency band, thus loosing representation in fine details. On the contrary, our method improves the performances without such side-effect.}
	\label{fig:frequency2}
	\vspace{-10pt}
\end{figure}


However, as \citet{liu2021discovering} point out, even trained with a large degradation pool, networks still have the tendency to overfit some specific degradations, effectively embedding degradation-related "semantics" within the network and causing reduced generalization. From this perspective, existing works that still adhere to straightforward optimization urgently need a proper training strategy (regularization) that helps to \textbf{make the best of the generalization knowledge hidden within the diverse degradations of training data.} Recently, \citet{kong2022reflash} make the first attempt to regularize networks with Dropout \cite{spatialdropout} and achieve appealing results. However, we argue that Dropout also brings side-effects that hurt high-frequency details in the restored image, leaving rooms for further improvements. Our method circumvents such problem by simply modulating feature statistics to encourage the model becomes indifferent to degradation-aware information, thus squeezing the last bit of degradation-invariant information in training data and improving generalization ability.

\begin{figure}[h]
	\centering
	\includegraphics[width=0.85\linewidth]{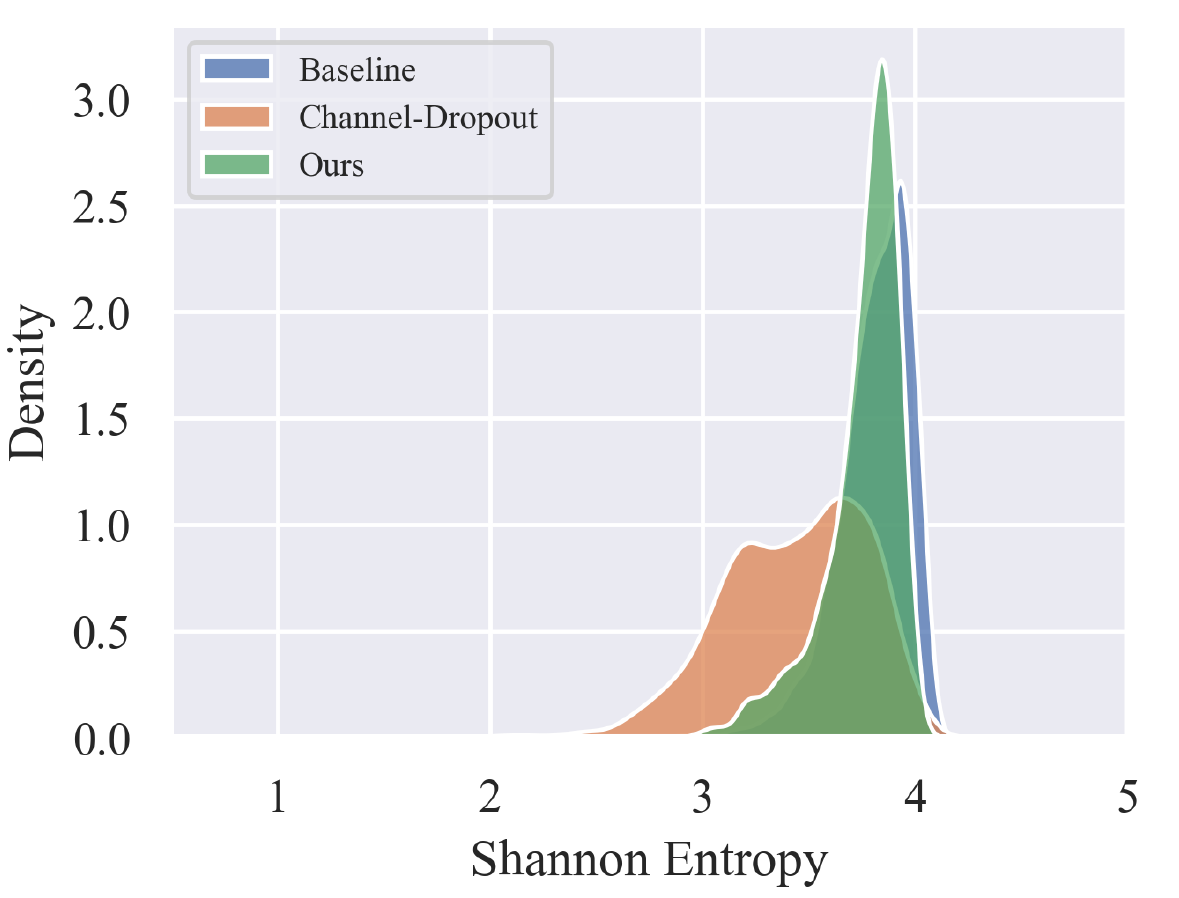}
	\vspace{-5pt}
	\caption{\textbf{Comparisons of the channel diversity from frequency perspective.} A higher entropy indicates a wider range of frequency bands covered by the model. Note that since it's hard to evaluate a pixel from frequency domain, we only consider from channel dimension and investigate the channel-wise dropout here.}
	\label{fig:channel_entropy}
	\vspace{-10pt}
\end{figure}

With much solid and inspiring work, researches based on the idea of the third brunch has gradually become the mainstream direction of recent Blind SR. Our method in fact serves as a complement to this line of work. In the future even with a larger and more realistic degradation pool, models trained in a straightforward manner still remain susceptible to the possibility of overfitting (e.g., some degradations are easier to learn than others), thus limiting their explorations to degradation-invariant representation (i.e., the ultimate goal of Blind SR). Therefore, our effort actually contributes to the research in a different way, neither from model design nor from building better dataset, but from proposing a training strategy (regularization) that could benefit both existing and future work.



\begin{figure*}[t]
	\centering
	\includegraphics[width=1\linewidth]{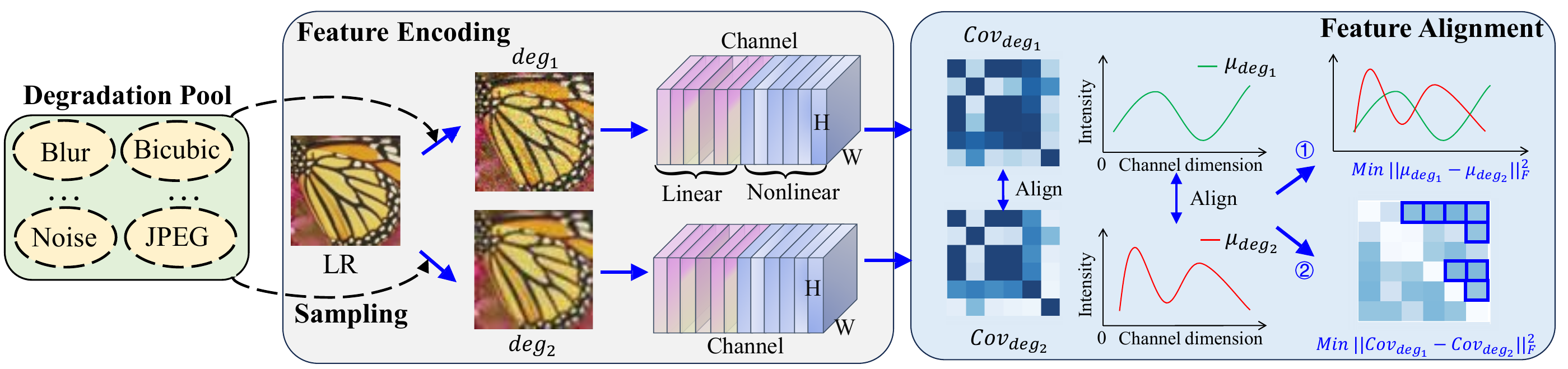}
	\caption{ \textbf{Overall schema of our proposed regularization.}
		Given two images with the same content but different degradations (e.g., blur and noise). We compute the mean and covariance of their features respectively (in both linear and nonlinear manner as we will elaborate in \secref{sec:experiment}). Then, a simple alignment is performed as regularization to encourage the model to learn more natural image prior without the disturbance of degradation information. In this way, the model will be more robust and generalizable against realistic unknown scenarios. }
	\label{fig:model}
\end{figure*}

\section{The Side-effect of Dropout}\label{sec:dropout}

Dropout \cite{srivastava2014dropout} and its variants \cite{spatialdropout,ghiasi2018dropblock} are fundamental techniques used in many high-level vision tasks (e.g., classification) to alleviate co-adaption and overfitting. However, every coin has its two sides, and the price for Dropout lies in the reduced feature interaction and diversity of the model. While this pose  nearly no threat to high-level vision tasks, it can severely impact the performances of image restoration. In this section, we first prove that Dropout decreases feature interactions, and then two experiments are conducted to support our point of view. 

Instead of considering the input variables working independently, DNNs encode the interaction between variables for better clues of inferences. For example, the restoration of a human face can be explained as
the interactions between left and right cheeks, between eyes and brows, etc. The interaction can be understood as follows. Still with the case of face restoration, let $\phi_{i=\text{left-cheek}}$ quantifies the numerical importance of the low-resolution left cheek region $i$ to its high-resolution counterpart when attempting to restore. Then, the interaction utility between the left cheek region $i$ and the right cheek region $j$ is measured as the change of $\phi_{i=\text{left-cheek}}$ value by the presence or absence of the right cheek region $j$. If the presence of $j$ in LR image increases $\phi_{i=\text{left-cheek}}$ by 0.1 (e.g., the left cheek can interact with right cheek to restore the similar shape and color), we then consider the utility of the interaction between $i$ and $j$ to be 0.1. 


We further extend the definition of interaction into multi-orders \cite{zhang2020game} to measure interactions of different complexities. Let \(N\) and \( S \) denote all input units and the context of interaction respectively, and \( T \subseteq S \) is a subset that indicates a specific pattern \( T \cup \{i,j\} \).  For any  given \( T \), \( R^T(i,j) \) quantifies the marginal reward obtained from the pattern of \( T \cup \{i,j\} \). Therefore we have the \( s \)-order interaction \( I^{(s)}(i,j) \) that measures the average interaction between variables \( (i,j) \) under all possible context with $s$ units as:
\begin{small}
	\begin{align*}
		I^{(s)}(i,j)&=\mathbb{E}_{S\subseteq N\setminus\{i,j\},|S|=s}\left[
		\sum\nolimits_{T\subseteq S}R^{T}(i,j)\right]
	\end{align*}
\end{small}
However, when the variables in context \( S \) are randomly removed by Dropout, the computation of \( I^{(s)}_\text{dropout}(i,j) \) will only involve the context in $S^\prime \subseteq S$ that are undropped: 
\begin{small}
	\begin{align*}
		I^{(s)}_\text{dropout}(i,j)&=
		\underset{S\subseteq N\setminus\{i,j\},|S|=s}{\mathbb{E}}
		\Big[
		\underset{S^\prime\subseteq S, |S^\prime|=r}{\mathbb{E}}
		\Big(\!\sum\nolimits_{T\subseteq S^\prime}R^T(i,j)\!\Big)\!\Big]
	\end{align*}
\end{small}

\begin{lemma}   When dropout is applied at a rate \( (1-p) \), the interaction \( I^{(s)}_{dropout}(i,j) \) only comprises of rewards from patterns with at most $r \sim B(s,p)$ units. Given this, \citet{zhang2020interpreting} proved that: $\frac{I_{\text {dropout }}^{(r)}(i, j)}{I^{(s)}(i, j)}=\frac{\sum\nolimits_{0\le q\le r}{\tbinom{r}{q}}J^{(q)}(i,j)}{\sum\nolimits_{0\le q\le s}{\tbinom{s}{q}}J^{(q)}(i,j)} \leq 1$, where \( J^q(i,j) = \mathbb{E}_{T \subseteq N \setminus \{i,j\}, |T|=q} [R^T(i,j)] \) is the average interaction for \((i,j)\) for all potential \(T\) with \(|T|=q\).  
\end{lemma}

The above derivation shows that when Dropout is applied to the model, the interaction of every order has become smaller. Extremely speaking, that means the left cheek region will never interact and gain information from the right cheek again, thus undermining its restored quality compared to models with high feature interaction. Note that the proof can also easily extend to channel dimension (i.e., channels work independently without interaction).

There are already many studies emphasizing the importance of interaction, both intra-channel \cite{chen2023better,bai2022improving,zhang2023pha,zhou2023fourmer} and inter-channel \cite{chen2023better,cui2023exploring,zamir2022restormer,SAN,cui2020color}, in the model's representation ability of high-frequency components (i.e., fine details such as edges and lines \cite{campbell1968application,de1980spatial}). Therefore, we conjecture that in SISR, applying Dropout will have negative impacts on fine-detail (high frequency) recovery, no matter it is the pixel-wise Dropout in \cite{srivastava2014dropout} or the channel-wise one used in \cite{kong2022reflash}. We show some simple observations here, and further provides the LPIPS result which is related to the high-frequency perceptual details \cite{ji2021non,chudasama2021rsrgan} in \tableref{tab:lpips} of \secref{sec:experiment}.




We visualize the error of the SRResNet model from the frequency perspective in \figref{fig:frequency2}. The model is trained on DIV2K with different strategies including the one proposed in this paper (see \secref{sec:method}), and tested on the six benchmark datasets with the settings of Real-ESRGAN \cite{ESRGAN} (i.e., Set5, Set14, BSD100, Test2k, Urban100, and Manga109). The error is estimated by transforming the images into frequency domain via Fast Fourier Transform (FFT). Then we introduce the Mean Absolute Percentage Error (MAPE) \cite{de2016mean} metric to indicate the error of each frequency band due to the imbalanced quantity between low and high frequency. Note that here larger MAPE values 
indicates greater error. 

From \figref{fig:frequency2}, we observe that as expected, model trained with Dropout has poorer performances in high frequency recovery. It's noteworthy that approximately 90\% of an image comprises low-frequency signals \cite{xu2022hinders}, and human perception is naturally sensitive to high-frequency details of an image. Thus, loosing the ability for high frequency restoration usually leads to unsatisfactory perception quality. 

Moreover, Dropout also tends to reduce feature diversity by smoothing out the activations of the network like a low-pass filter \cite{zhang2019confidence}. Nevertheless, it is acknowledged that the diversity of features actually contributes to the representation ability of different frequency information \cite{qin2021fcanet,Magid_2021_ICCV}. In the case of SR, as previously mentioned, low-frequency signals dominate the natural images. Therefore while the network trained with Dropout has no choice but to concentrate the representation power of features into low-frequency (i.e., the features are not diverse enough to represent a wide range of frequency information), network trained without Dropout enjoys diverse feature representation ability in different frequencies. We conduct an auxiliary experiment in \figref{fig:channel_entropy}, which uses Discrete Cosine Transform 
as in \cite{qin2021fcanet} to first identify the representative frequency band for each channel, and then computes the Shannon Entropy across channels to reveal the range of frequency band information encoded by the model. The experiment is run with SRResNet on the previously mentioned six datasets, and we present the averaged results here. Without surprise, we observe that model trained with Dropout covers a smaller range of frequency band, limiting its restoration power outside this range.

\section{Simple Alignments as Regularization }\label{sec:method}

We revealed the drawbacks of applying Dropout in SR (i.e., reducing feature interactions and diversity) in the previous section. In this section, we will show how a simple alignment can effectively improve the performance of Blind SR. The overall schema of our method is shown in \figref{fig:model}.

The idea of our method stems from the aspiration that model should make predictions independent of different degradations. For example, given two images with the same content but different degradations \(deg_1\) and \(deg_2\), the model is expected to output the same restored image \(I_{o}\) from these two inputs, \textit{i.e.,} \(P(I_o | I_d, d = deg_1) = P(I_o | I_d, d = deg_2)\). While this seems straightforward by just forcing their intermediate features to be exactly the same, we argue that it will be too harsh and overly constrain the model, hindering its ability to reach a local minimum effectively (\textit{An ablation study showing its inferior performances is in supplemental material}). Instead, in this paper we draw inspirations from image style transfer \cite{huang2017arbitrary,ulyanov2017improved,li2017demystifying} and treat images with different degradations as with different styles. Note that the similar idea is also adopted in \cite{li2022face}, but they are focusing on degradation generation and thus not in our discussion scope. 

Then, we follow the traditions of style transfer and utilize the mean and covariance as the degradation (style) sensitive indicators \cite{huang2017arbitrary}. Although such choice seems lacking solid theoretical foundation, it actually aligns with the research instincts: mean and covariance are two commonly used first and second order statistics in image processing, and have been shown to reflect the global status of  activations and the detailed structure and texture respectively \cite{gao2021representative,SAN,karacan2013structure}. We hypothesize that different degradations should have different impacts on these aspects, therefore making the use of these two statistics reasonable. Empirical studies in \secref{sec:experiment} also provide strong evidence for their effectiveness. Consequently, by aligning these statistics across images that have the same content but different degradations, we aim to guide the model to 
automatically ignore the information specific to degradation during feature encoding, thereby improving its learning ability of degradation-invariant features that are crucially needed for handling new and unknown degradations. Next, we show how exactly the alignment is carried out in both linear and nonlinear manners.


\noindent\textbf{Linear Alignment} is in its most imaginable form which punishes the statistical differences with Frobenius norm. Given features $\boldsymbol{x}$ and $\boldsymbol{x}^{\prime}$ from images with varying degradation but identical content, the regularization is defined as:
\begin{equation}
	\begin{aligned}
		{\ell_{lin.}}= {\| Cov(\boldsymbol{x}) - Cov(\boldsymbol{x}^{\prime}) \|}^2_F + \|\mu(\boldsymbol{x}) - \mu(\boldsymbol{x}^{\prime}) \|^2_F, \nonumber
	\end{aligned}
	\label{eq:coral}
\end{equation}
where ${\|\cdot\|}^2_F$ represents the squared matrix Frobenius norm. 
The mean  $\mu(\boldsymbol{z}) = \frac{1}{hw} \sum_i \sum_j \boldsymbol{z}_{:, i, j} \label{eq:cov_s}$ and covariance $Cov$ is
$Cov(\boldsymbol{z})= {\frac{1}{C-1}}({\boldsymbol{z}^{\top} \boldsymbol{z} - \frac{1}{C}{({\mathbbm{1}}^{\top}\boldsymbol{z}})^{\top} {({\mathbbm{1}}^{\top}\boldsymbol{z}})})$.


\begin{table*}[ht!]
	\setlength{\abovecaptionskip}{-2pt}
	\setlength{\belowcaptionskip}{-4pt}
	\renewcommand{\arraystretch}{0.9}
	\caption{\textbf{Six benchmarks with eight types of degradations (clean, noise, blur, jpeg, blur+noise, blur+jpeg, noise+jpeg, and blur+noise+jpeg) are used to evaluate the PSNR (dB) results in $\times$4 resolution settings. Degradations are abbreviated in table.}}
	
	\begin{center}
		\resizebox{\linewidth}{!}{
			\begin{tabular}{|l|cccc|cccc|cccc|}
				\hline
				\multirow{2}{*}{Models}                 & \multicolumn{4}{c|}{Set5~\cite{Set5}}                                                        & \multicolumn{4}{c|}{Set14~\cite{Set14}}                                                       & \multicolumn{4}{c|}{BSD100~\cite{BSD100}}                                                                                                      \\ \cline{2-13}
				&             \multicolumn{1}{c}{clean}                        & blur                         & \multicolumn{1}{c}{noise}                        & jpeg                         &  \multicolumn{1}{c}{clean}                        & blur                         & \multicolumn{1}{c}{noise}                        & jpeg                         & \multicolumn{1}{c}{clean}                     & blur            
				& \multicolumn{1}{c}{noise}                        & jpeg                 
				\\ \hline
				SRResNet \cite{SRResNet}         &       24.85&24.73&22.52&23.67&23.25&23.05&21.18&22.32&23.06&22.99&21.34&22.47 \\ 
				+ Ours         &  25.93&25.62&23.15&24.38&24.12&23.80&21.67&22.99& 23.83&23.64&21.77&23.04\\ 
				\textcolor{red}{Improvement}                  &     \textcolor{red}{+1.08}&\textcolor{red}{+0.89}&\textcolor{red}{+0.63}&\textcolor{red}{+0.71}&\textcolor{red}{+0.87}&\textcolor{red}{+0.75}&\textcolor{red}{+0.49}&\textcolor{red}{+0.67}&\textcolor{red}{+0.77}&\textcolor{red}{+0.65}&\textcolor{red}{+0.43}&\textcolor{red}{+0.57}   \\ \hline
				RRDB    \cite{ESRGAN}          &     25.18&25.12&21.79&23.82&23.74&23.36&21.02&22.59&23.38&23.32&21.00&22.73     \\ 
				+ Ours            &          26.78&26.55&23.02&24.70&24.70&24.35&21.91&23.21& 24.59& 24.54& 23.47& 23.67       \\ 
				\textcolor{red}{Improvement}                      &  \textcolor{red}{+1.60}&\textcolor{red}{+1.43}&\textcolor{red}{+1.23}&\textcolor{red}{+0.88}&\textcolor{red}{+0.96}&\textcolor{red}{+0.99}&\textcolor{red}{+0.89}&\textcolor{red}{+0.62}&\textcolor{red}{+1.21}&\textcolor{red}{+1.22}&\textcolor{red}{+2.47}&\textcolor{red}{+0.94}
				\\ 
				\hline
				MSRN     \cite{li2018multi}           &   25.25&24.89&22.57&24.08&23.38&23.10&21.80&22.53&23.38&23.30&21.92&22.76\\ 
				+ Ours            &  25.81&25.52&22.84&24.46&23.93&23.64&21.86&22.83&23.72&23.58&22.01&22.98        \\ 
				\textcolor{red}{Improvement}                      &    \textcolor{red}{+0.56}&\textcolor{red}{+0.63}&\textcolor{red}{+0.27}&\textcolor{red}{+0.38}&\textcolor{red}{+0.55}&\textcolor{red}{+0.54}&\textcolor{red}{+0.06}&\textcolor{red}{+0.30}&\textcolor{red}{+0.34}&\textcolor{red}{+0.28}&\textcolor{red}{+0.09}&\textcolor{red}{+0.22} \\ \hline
				SwinIR     \cite{liang2021swinir}           &        26.25&26.03&22.96&24.37&24.53&24.25&22.08&23.14&23.91&23.83&22.12&23.04  \\ 
				+ Ours            &         26.49&26.23&24.61&24.68&24.65&24.28&22.23&23.29&24.04&23.96&22.21&23.15  \\ 
				\textcolor{red}{Improvement}                      &    \textcolor{red}{+0.24}&\textcolor{red}{+0.20}&\textcolor{red}{+1.65}&\textcolor{red}{+0.31}&\textcolor{red}{+0.12}&\textcolor{red}{+0.03}&\textcolor{red}{+0.15}&\textcolor{red}{+0.15}&\textcolor{red}{+0.13}&\textcolor{red}{+0.13}&\textcolor{red}{+0.09}&\textcolor{red}{+0.11}  \\ 
				\toprule\toprule
				& b+n                     & b+j                         & n+j                        & b+n+j                       & b+n                     & b+j                         & n+j                        & b+n+j    & b+n                     & b+j                         & n+j                        & b+n+j                \\ \hline
				
				SRResNet \cite{SRResNet}         &   23.27&23.40&23.05&22.73&22.23&22.06&21.99&21.77&22.25&22.33&22.22&22.04     \\ 
				+ Ours         
				&23.79&23.86&23.71&23.19 &22.65&22.63&22.55&22.16&22.53&22.79&22.62&22.32\\ 
				\textcolor{red}{Improvement}                  &   \textcolor{red}{+0.52}&\textcolor{red}{+0.46}&\textcolor{red}{+0.66}&\textcolor{red}{+0.46}&\textcolor{red}{+0.42}&\textcolor{red}{+0.57}&\textcolor{red}{+0.56}&\textcolor{red}{+0.39}&\textcolor{red}{+0.28}&\textcolor{red}{+0.46}&\textcolor{red}{+0.40}&\textcolor{red}{+0.28}     \\ \hline
				RRDB    \cite{ESRGAN}          &    23.44&23.45&23.32&22.81&22.47&22.17&22.29&21.95&22.39&22.47&22.42&22.15   \\ 
				+ Ours            &   24.12&24.14&23.93&23.26&22.80&22.76&22.71&22.21& 22.85& 23.21& 22.97& 22.54            \\ 
				\textcolor{red}{Improvement}                      &   \textcolor{red}{+0.68}&\textcolor{red}{+0.69}&\textcolor{red}{+0.61}&\textcolor{red}{+0.45}&\textcolor{red}{+0.33}&\textcolor{red}{+0.59}&\textcolor{red}{+0.42}&\textcolor{red}{+0.26}&\textcolor{red}{+0.46}&\textcolor{red}{+0.74}&\textcolor{red}{+0.55}&\textcolor{red}{+0.39}   \\ \hline
				MSRN     \cite{li2018multi}           &   23.55&23.59&23.50&22.95&22.39&22.23&22.19&21.97&22.57&22.61&22.45&22.24   \\ 
				+ Ours            & 23.70&23.80&23.73&23.06&22.52&22.49&22.48&22.08&22.68&22.73&22.56&22.26           \\ 
				\textcolor{red}{Improvement}                      &    \textcolor{red}{+0.15}&\textcolor{red}{+0.21}&\textcolor{red}{+0.23}&\textcolor{red}{+0.11}&\textcolor{red}{+0.13}&\textcolor{red}{+0.26}&\textcolor{red}{+0.29}&\textcolor{red}{+0.11}&\textcolor{red}{+0.11}&\textcolor{red}{+0.12}&\textcolor{red}{+0.11}&\textcolor{red}{+0.02}   \\ \hline
				SwinIR     \cite{liang2021swinir}           &  23.80&23.84&23.67&22.99&22.53&22.73&22.59&22.20&22.61&22.82&22.61&22.34      \\ 
				+ Ours            &       24.13&24.17&23.89&23.09&22.87&22.79&22.81&22.28&22.77&22.98&22.76&22.40        \\ 
				\textcolor{red}{Improvement}                      &     \textcolor{red}{+0.33}&\textcolor{red}{+0.33}&\textcolor{red}{+0.22}&\textcolor{red}{+0.10}&\textcolor{red}{+0.34}&\textcolor{red}{+0.06}&\textcolor{red}{+0.22}&\textcolor{red}{+0.08}&\textcolor{red}{+0.16}&\textcolor{red}{+0.16}&\textcolor{red}{+0.15}&\textcolor{red}{+0.06}  \\ \bottomrule
				\toprule
				\multirow{2}{*}{Models}                 & \multicolumn{4}{c|}{Test2k~\cite{kong2021classsr}}                                                        & \multicolumn{4}{c|}{Urban100~\cite{Urban100}}                                                       & \multicolumn{4}{c|}{Manga109~\cite{Manga109}}                                                                                                      \\ \cline{2-13}
				&             \multicolumn{1}{c}{clean}                        & blur                         & \multicolumn{1}{c}{noise}                        & jpeg                         &  \multicolumn{1}{c}{clean}                        & blur                         & \multicolumn{1}{c}{noise}                        & jpeg                         & \multicolumn{1}{c}{clean}                     & blur            
				& \multicolumn{1}{c}{noise}                        & jpeg                 
				\\ \hline
				SRResNet \cite{SRResNet}         &     23.91&23.71&21.77&23.11&21.23&21.06&19.74&20.60&18.42&18.75&18.08&18.27 \\ 
				+ Ours         &  24.58&24.43&22.17&23.65&21.94&21.65&20.19&21.20 &19.18&19.46&18.90&19.02\\ 
				\textcolor{red}{Improvement}                  &     \textcolor{red}{+0.67}&\textcolor{red}{+0.72}&\textcolor{red}{+0.40}&\textcolor{red}{+0.54}&\textcolor{red}{+0.71}&\textcolor{red}{+0.59}&\textcolor{red}{+0.45}&\textcolor{red}{+0.60}&\textcolor{red}{+0.76}&\textcolor{red}{+0.71}&\textcolor{red}{+0.82}&\textcolor{red}{+0.75}   \\ \hline
				RRDB    \cite{ESRGAN}          &     24.16&23.64&21.34&23.36 &21.57&21.18&19.61&20.93&18.59&18.64&18.30&18.41 \\ 
				+ Ours            & 24.97&24.76&22.15&23.86&22.29&21.95&20.21&21.40&19.40&19.61&18.96&19.24                 \\ 
				\textcolor{red}{Improvement}                      &     \textcolor{red}{+0.81}&\textcolor{red}{+1.12}&\textcolor{red}{+0.81}&\textcolor{red}{+0.50}&\textcolor{red}{+0.72}&\textcolor{red}{+0.77}&\textcolor{red}{+0.60}&\textcolor{red}{+0.47}&\textcolor{red}{+0.81}&\textcolor{red}{+0.97}&\textcolor{red}{+0.66}&\textcolor{red}{+0.83}  \\ \hline
				MSRN     \cite{li2018multi}           &   22.99& 23.83 &22.30&23.22 &21.35&21.14&20.19&20.75&19.12&19.31&18.72&18.89\\ 
				+ Ours            &    24.52&24.23&22.38&23.56&21.88&21.54&20.22&21.14&19.23&19.35&18.84&19.01     \\ 
				\textcolor{red}{Improvement}                      &      \textcolor{red}{+1.53}&\textcolor{red}{+0.40}&\textcolor{red}{+0.08}&\textcolor{red}{+0.34}&\textcolor{red}{+0.53}&\textcolor{red}{+0.40}&\textcolor{red}{+0.03}&\textcolor{red}{+0.39}&\textcolor{red}{+0.11}&\textcolor{red}{+0.04}&\textcolor{red}{+0.12}&\textcolor{red}{+0.12}  \\ \hline
				SwinIR     \cite{liang2021swinir}           &        24.78 &24.57&22.71&23.63&22.18&21.90&20.56&21.32&19.10&19.27&18.71&18.95 \\ 
				+ Ours            &  24.98&24.76&22.84&23.80&22.34&22.07&20.69&21.48&19.24&19.45&18.98&19.28         \\ 
				\textcolor{red}{Improvement}                      &   \textcolor{red}{+0.20}&\textcolor{red}{+0.19}&\textcolor{red}{+0.13}&\textcolor{red}{+0.17}&\textcolor{red}{+0.16}&\textcolor{red}{+0.17}&\textcolor{red}{+0.13}&\textcolor{red}{+0.16}&\textcolor{red}{+0.14}&\textcolor{red}{+0.18}&\textcolor{red}{+0.27}&\textcolor{red}{+0.33}    \\ 
				\toprule\toprule
				& b+n                     & b+j                         & n+j                        & b+n+j                       & b+n                     & b+j                         & n+j                        & b+n+j    & b+n                     & b+j                         & n+j                        & b+n+j                      \\ \hline
				SRResNet \cite{SRResNet}         &   22.81&22.87&22.85&22.59  &20.46&20.30&20.42&20.09&18.59&18.50&18.21&18.39 \\ 
				+ Ours         &  23.11&23.27&23.22&22.82&20.73&20.72&20.91&20.37&19.27&19.17&18.98&19.01\\ 
				\textcolor{red}{Improvement}                  &   \textcolor{red}{+0.30}&\textcolor{red}{+0.40}&\textcolor{red}{+0.37}&\textcolor{red}{+0.23}&\textcolor{red}{+0.27}&\textcolor{red}{+0.42}&\textcolor{red}{+0.49}&\textcolor{red}{+0.28}&\textcolor{red}{+0.68}&\textcolor{red}{+0.67}&\textcolor{red}{+0.77}&\textcolor{red}{+0.62}     \\ \hline
				RRDB    \cite{ESRGAN}          &    22.93&22.87&23.12&22.73 &20.57&20.40&20.74&20.24&18.83&18.43&18.38&18.41 \\ 
				+ Ours            &   23.14&23.37&23.34&22.82&20.76&20.85&21.03&20.38&19.43&19.31&19.12&19.15      \\ 
				\textcolor{red}{Improvement}                      &    \textcolor{red}{+0.21}&\textcolor{red}{+0.50}&\textcolor{red}{+0.22}&\textcolor{red}{+0.09}&\textcolor{red}{+0.19}&\textcolor{red}{+0.45}&\textcolor{red}{+0.29}&\textcolor{red}{+0.14}&\textcolor{red}{+0.60}&\textcolor{red}{+0.88}&\textcolor{red}{+0.74}&\textcolor{red}{+0.74}  \\ \hline
				MSRN     \cite{li2018multi}           &   23.03&23.01&22.94&22.66 &20.65&20.43&20.56&20.19&19.16&19.02&18.80&18.88  \\ 
				+ Ours      & 23.21&23.21&23.18&22.78&20.76&20.64&20.89&20.26&19.19&19.18&18.92&18.93              \\ 
				\textcolor{red}{Improvement}                      &     \textcolor{red}{+0.18}&\textcolor{red}{+0.20}&\textcolor{red}{+0.24}&\textcolor{red}{+0.12}&\textcolor{red}{+0.11}&\textcolor{red}{+0.21}&\textcolor{red}{+0.33}&\textcolor{red}{+0.07}&\textcolor{red}{+0.03}&\textcolor{red}{+0.16}&\textcolor{red}{+0.12}&\textcolor{red}{+0.05}  \\ \hline
				SwinIR     \cite{liang2021swinir}           &   23.15&23.27&23.21&22.81 &20.89&20.79&20.98&20.45&19.07&19.02&18.79&18.80   \\ 
				+ Ours            &     23.35&23.47&23.45&22.93&21.02&20.98&21.12&20.53&19.37&19.35&19.15&19.12          \\ 
				\textcolor{red}{Improvement}                      &  \textcolor{red}{+0.20}&\textcolor{red}{+0.20}&\textcolor{red}{+0.24}&\textcolor{red}{+0.12}&\textcolor{red}{+0.13}&\textcolor{red}{+0.19}&\textcolor{red}{+0.14}&\textcolor{red}{+0.08}&\textcolor{red}{+0.30}&\textcolor{red}{+0.33}&\textcolor{red}{+0.36}&\textcolor{red}{+0.32}     \\ \hline
		\end{tabular}}
	\end{center}

	\label{table:degradations1}
	\vskip -0.30cm
\end{table*}

\noindent\textbf{Nonlinear Alignment} can be considered as an enhanced version of the linear alignment to overcome the dimensional constraints of encoder models in SR. These models, due to practicality, cannot be arbitrarily scaled up. However theoretically, as feature dimension increases, distributions indistinguishable in low dimensions might become separable in higher ones (i.e., nonlinear property can gradually becomes linearly observable in higher dimensions) \cite{bhavsar2012review}. Therefore, we would like to extend the alignment beyond the dimensional limit of the model to align the distributions of different degradations both within the dimensional capacity of the model and beyond it in higher or even infinite dimension. We believe the knowledge glimpsed from the higher dimension can effectively ``highlight" the deeper differences between degradations and thus, serves as a ``look-ahead" to further steer the model towards degradation-invariant. Meanwhile, the gradients from higher-dimensional alignment might also act like a perceptive teacher to facilitate the optimization of alignment in lower dimensions. 



To this end, in order to peep secrets from the higher dimension in a parameter-efficient way, let's consider a feature mapping $\Phi: \mathcal{X} \rightarrow \mathcal{H}$, which transforms elements in $\mathcal{X}$ into a reproducing kernel Hilbert space (RKHS) denoted as $\mathcal{H}$. Instead of derivating the explicit expression of $\Phi$, in this paper we utilize the Random Fourier features (RFF) \cite{rahimi2007random} to approximate the behavior of Radial Basis Function (RBF) kernel to explicitly project features into $\mathcal{H}_{RFF}$ where data has limited dimension but shows similar properties as if in infinite dimension space. Therefore, let $\boldsymbol{u}$ and $\boldsymbol{v}$ be the RFF mapping function satisfying $\boldsymbol{u}(z), \boldsymbol{v}(z)\in \mathcal{H}_{\text{RFF}}$, we have our nonlinear alignment as:
\begin{small}
	\begin{align*}
		&\ell_{nonlin.} = \| Cov(\boldsymbol{h}_{\boldsymbol{x}}) - Cov(\boldsymbol{h}_{\boldsymbol{x}^{\prime}}) \|_F^2 
		+ \|\mu(\boldsymbol{h}_{\boldsymbol{x}}) - \mu(\boldsymbol{v}(\boldsymbol{h}_{\boldsymbol{x}^{\prime}}) \|_F^2,  \\
		&\boldsymbol{h}_{\boldsymbol{x}} = \left(\boldsymbol{u}^1(\boldsymbol{x}), \dots, \boldsymbol{u}^n(\boldsymbol{x})\right),  \quad
		\boldsymbol{h}_{\boldsymbol{x}^{\prime}}  = \left(\boldsymbol{v}^1(\boldsymbol{x}^\prime), \dots, \boldsymbol{v}^n(\boldsymbol{x}^\prime)\right), \\
		&\mathcal{H}_{\text{RFF}}= \left\{ h \colon  \sqrt{2}\cos(\omega x + \phi) \mid  \omega \sim \mathcal{N}(0, 1), \phi \sim U(0, 2\pi) \right\}.
	\end{align*}  
\end{small}

\begin{figure*}[t]
	\centering
	\includegraphics[width=1\linewidth]{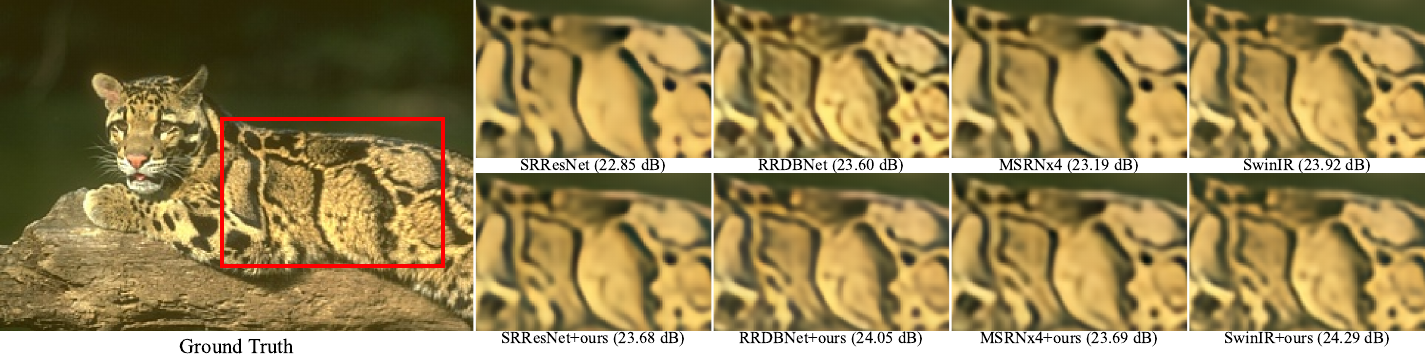}\\
	\includegraphics[width=1\linewidth]{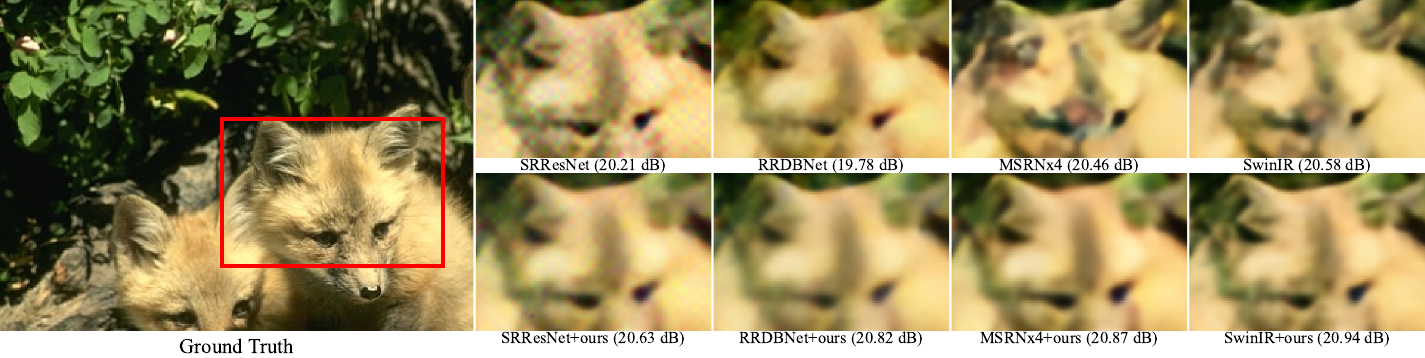}
	\includegraphics[width=1\linewidth]{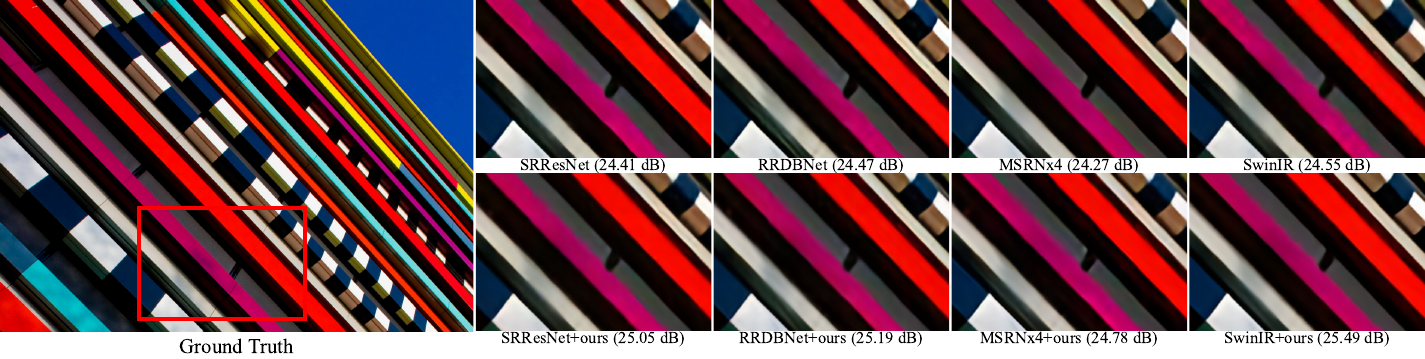}
	\caption{\textbf{Visual results of representative degradations (blur, noise, and JPEG) on baseline models w/ and w/o our regularization.}}
	\label{fig:compare}
	\vspace{-5pt}
\end{figure*}

\noindent\textbf{Discussion.} Thanks to the advancement of realistic stochastic degradation generation methods, training models with multi-degradations has become standard for most recent Blind SR works. Our method integrates seamlessly with them by simply requiring the degradation generation model (e.g., \cite{wang2021realesrgan,luo2022learning}) to randomly generate one more degradation for each (or some) image(s) to form the ($\boldsymbol{x}$,$\boldsymbol{x}^{\prime}$) pairs for regularization. Note that the forward pass of ($\boldsymbol{x}$,$\boldsymbol{x}^{\prime}$) can be efficiently parallelized across multiple GPUs. Furthermore, explicitly aligning only the feature's first and second order statistics also makes our method reliable and efficient.

\section{Experiment}\label{sec:experiment}
In this paper we follow \citet{kong2022reflash} and adopt the widely acknowledged multi-degradations settings used in Blind SR researches \cite{wang2021realesrgan} to make fair and credible comparisons. \textit{Due to space limits, we refer readers to our supplementary materials for more experimental settings and ablation studies.}

\begin{figure}[h]
\centering
\subfloat[SRResNet]{\includegraphics[width=0.48\linewidth]{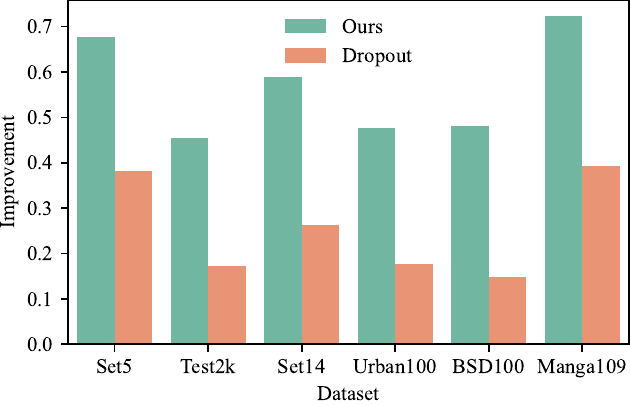}}
\hspace*{\fill}
\subfloat[RRDB]{\includegraphics[width=0.48\linewidth]{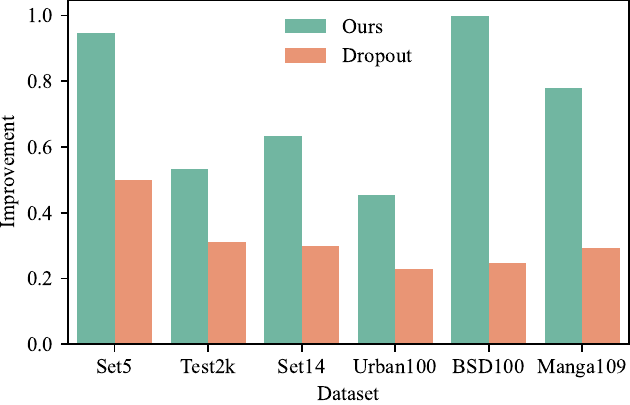}}

\subfloat[MSRN]{\includegraphics[width=0.48\linewidth]{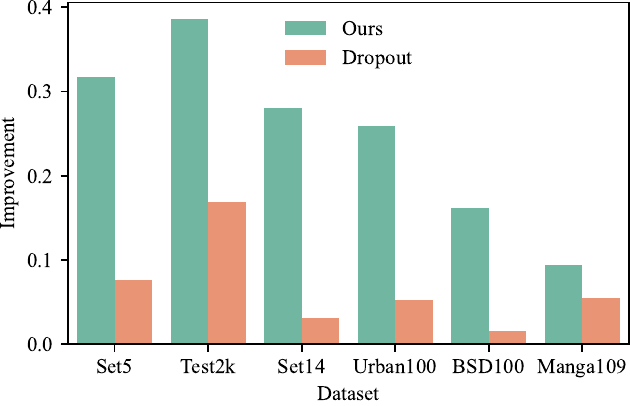}}
\hspace*{\fill}
\subfloat[SwinIR]{\includegraphics[width=0.48\linewidth]{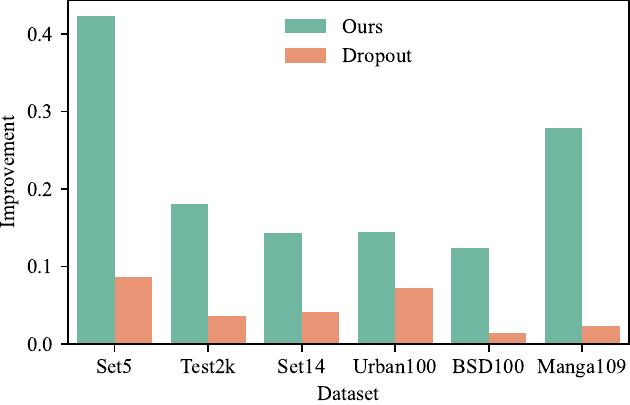}}

\caption{\textbf{Average improvements on baseline models of our method and Dropout on six benchmark datasets. }}
\label{fig:compare_drop2}
\end{figure}

\begin{table*}[!t]
\setlength{\belowcaptionskip}{-1pt}
\renewcommand{\arraystretch}{1.1}
\caption{\textbf{Comparisons with Dropout using both PSNR (run on realistic data of NTIRE 2018 challenge) and LPIPS (run on six widely used benchmarks).}}\label{tab:lpips}
\centering
\setlength{\tabcolsep}{1.0mm}{
	\resizebox{0.7\linewidth}{!}{%
		\begin{tabular}{|l|c | c|}
			\hline
			\multirow{2}*{Models} & \multicolumn{1}{c|}{PSNR $\uparrow$} & \multicolumn{1}{c|}{LPIPS $\downarrow$} \\
			\cline{2-3}
			& mild / difficult / wild & Set5 / Set14 / BSD / Manga / Urban / Test2k \\
			\hline
			SRResNet & 16.94 / 17.84 / 17.55 & 0.241 / 0.353 / 0.433 / 0.220 / 0.323 / 0.369 \\
			+Dropout & 17.10 / 18.02 / 17.78 & 0.245 / 0.356 / 0.439 / 0.223 / 0.333 / 0.374 \\
			+\textit{Ours} & \textbf{17.69} / \textbf{18.30} / \textbf{17.94} & \textbf{0.238} / \textbf{0.349} / \textbf{0.429} / \textbf{0.219} / \textbf{0.319}  / \textbf{0.367} \\
			\hline
			RRDB & 16.79 / 17.66 / 17.38 & 0.218 / 0.324 / 0.398 / 0.206 / 0.300 / 0.344 \\
			+Dropout & 17.31 / 18.17 / 17.89 & 0.229 / 0.334 / 0.412 / 0.210 / 0.309 / 0.354 \\
			+\textit{Ours} & \textbf{17.88} / \textbf{18.47} / \textbf{18.21} & \textbf{0.217} / \textbf{0.322} / \textbf{0.397} / \textbf{0.205} / \textbf{0.299} / \textbf{0.341} \\
			\hline
		\end{tabular}%
	}
}
\end{table*}

\begin{figure}[t]
\centering
\includegraphics[width=0.48\linewidth]{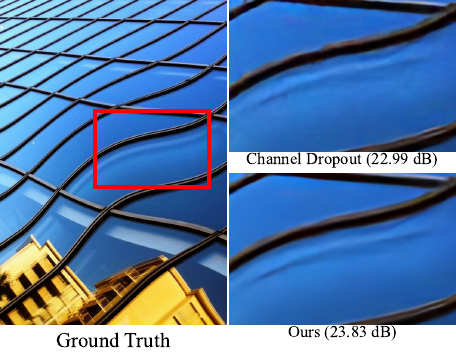}
\includegraphics[width=0.48\linewidth]{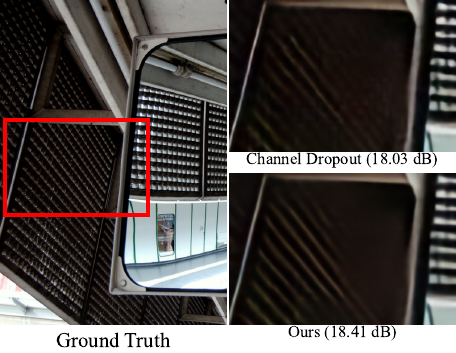}

\includegraphics[width=0.48\linewidth]{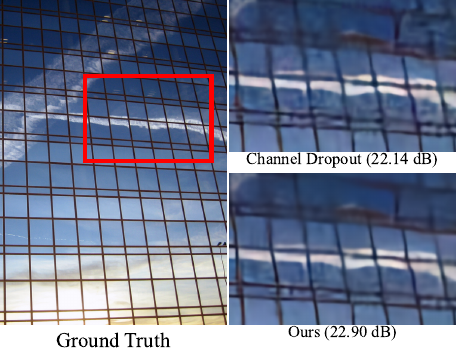}
\includegraphics[width=0.48\linewidth]{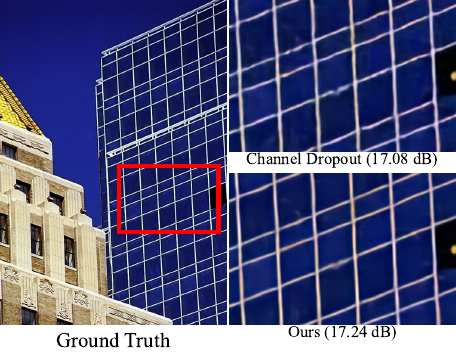}
\caption{\textbf{Restoration results of Dropout and our method.}}
\label{fig:drop_compare}
\end{figure}

\noindent \textbf{Improvements on Baseline Models.}
We show the improvements of applying our methods on the top of baseline models in \tableref{table:degradations1}. Notably, we achieve considerable performance improvements in almost all cases (averaged 0.44dB). We draw three conclusions from \tableref{table:degradations1}: (1) the test case of blur + noise + jpeg (b+n+j) has universally the smallest improvement (averaged 0.21dB) to baseline models compared with other degradations types (averaged 0.45dB). We hypothesize that this is because b+n+j is actually an ``in-distribution" test setting where the models already have good enough performances. We call it ``in-distribution" because following \citet{wang2021realesrgan}, the model is also (and only) trained with such setting, although with different degradation parameters. Since our method is designed to improve performances over unknown degradations, it is not surprising that we don't have impressive results with ``in-distribution" setting. (2) On the contrary, the test cases of clean, blur, noise and jpeg, whose patterns deviate the most from training distribution (i.e., b+n+j), have shown excellent improvements (averaged 0.55dB) compared with other settings (averaged 0.31dB), demonstrating the improved generalization ability offered by our method. (3) From the perspective of models, SRResNet and RRDB show more improvements (averaged 0.65dB) than MSRN and SwinIR (averaged 0.22dB). We speculate that it's because MSRN and SwinIR introduce extra multi-scale and transformer architecture over the residual CNNs of SRResNet and RRDB, suggesting that multi-scale feature and larger receptive field might also benefit model generalization in Blind SR. We also show the visual comparisons of baseline models w/ and w/o our regularization in \figref{fig:compare}, and as observed, models regularized by our method generally perform better in content reconstruction and artifact removal.

\noindent \textbf{Performance Comparison with Dropout.}
The comparison of average improvement between Dropout (unless otherwise specified, Dropout used in our experiments refers to the channel-wise Dropout used in \cite{kong2022reflash}) and our method is illustrated in 
\figref{fig:compare_drop2}. It is noteworthy that our method outperforms Dropout on all models and datasets, with an average improvement of 0.57dB v.s. 0.26dB on SRResNet,  0.72dB v.s. 0.31dB  on RRDB, 0.25dB v.s. 0.07dB  on MSRN and 0.22dB v.s. 0.05dB on SwinIR. We argue that the performance gap observed here is reasonable, because our method in essence explicitly aligns different degradations to encourage the model becomes degradation-invariant. However, Dropout, aside from its side-effects discussed earlier, only guides such invariant feature learning in a rather implicit manner, and thus can't thoroughly get rid of degradation-specific information, leading to inferior generalization ability. \textit{The detailed data of \figref{fig:compare_drop2} is in supplementary material}.

\begin{figure}[t]
\centering
\includegraphics[width=1\linewidth]{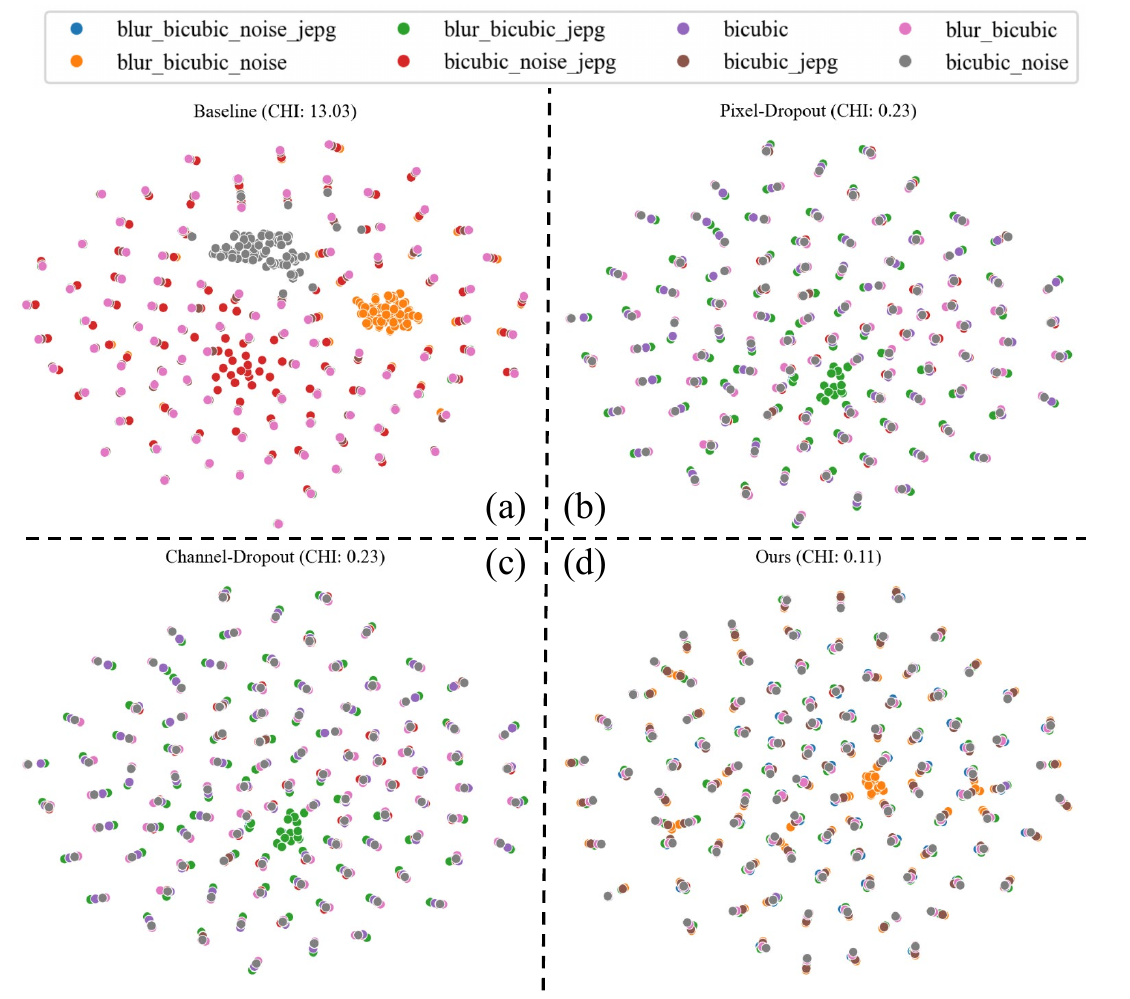}
\caption{\textbf{The visualization of the DDR clusters of SRResNet trained with different regularizations. The CHI results are also provided to measure the separation degree of clusters.}}
\label{fig:chi}
\end{figure}


Although the multi-degradation setting of \cite{wang2021realesrgan} has been proven effective in stimulating real-world degradations, we further evaluate our approach against Dropout with realistic NTIRE 2018 SR challenge data \cite{timofte2017ntire}. The results are presented in the left part of \tableref{tab:lpips} with our method  showing overwhelming advantages in generalization over Dropout. Moreover, we also investigate the model performances using the LPIPS metric \cite{zhang2018unreasonable} which is more relevant to perceptual details of an image in the right part of \tableref{tab:lpips}. The results of LPIPS correspond to our previous discussion that while Dropout improves PSNR by alleviating overfitting, it could simultaneously bring negative impacts on the recovery of fine-grained details (reflecting by LPIPS). On the contrary, our method can enjoy the best of both worlds without such side-effect. We also visualize the restored images of Dropout and our method for vivid comparison in \figref{fig:drop_compare}. 


Following \citet{kong2022reflash}, we also adopt the deep degradation representation (DDR) introduced by \cite{liu2021discovering} and visualized it in \figref{fig:chi}. In the figure, each point represents an input image and different colors indicate different degradations. DDR provides us a way to assess the network's generalization ability by peeking into the model behaviors. For example, in \figref{fig:chi} (a) we can observe that images with the same degradations are clustered together, which means the model has learned to encode degradation-specific information, leading to its poor generalization ability. On the other hand, in \figref{fig:chi} (d) images are clustered relying more on their contents instead of degradations, which means the model has become more degradation-invariant. \citet{liu2021discovering} further introduce the Calinski-Harabaz Index (CHI) \cite{calinski1974dendrite} for quantitative analysis, with a lower value indicating better cluster separation, and thus better generalization ability.

\section{Conclusion}
From theoretical and experimental aspects, this paper first reveal the side-effects of applying Dropout as a regularizer in SR. Then as an alternative, we propose a simple and effective feature alignment regularization that can further enhance the generalization ability for Blind SR models. \textit{Given the current challenges in advancing Blind SR researches, we call for more efforts exploring training regularization, a path not fully developed but potentially highly impactful.}

\section{Acknowledgments}
This work was supported by JSPS KAKENHI Grant Numbers 22H00529, 20H05951, JST-Mirai Program JPMJMI23G1 and JST SPRING, Grant Number JPMJSP2108.



{
\small
\bibliographystyle{ieeenat_fullname}
\bibliography{egbib}
}


\clearpage
\setcounter{page}{1}
\setcounter{figure}{0}
\setcounter{section}{0}
\setcounter{table}{0} 
\maketitlesupplementary
\section{Experiment Settings} 

In order to simulate real-world degradations better, most state-of-the-art Blind SR researches examine their methods with the multi-degradations settings. However, since there is no unified standards for how the multi-degradations should be generated, different works usually employ it in their own ways. In this paper, for the purpose of a fair and credible validation of our method, we choose the widely adopted ``second-order" degradation generation settings of \citet{wang2021realesrgan} to verify our effectiveness for Blind SR. Note that the Dropout \cite{kong2022reflash}, which will be compared with our method in experiments, also adopt the same setting.

For our training, we leverage the high-resolution (HR) images from the DIV2K~\cite{DIV2K} dataset. During the training process, the L1 loss function is employed in combination with the Adam optimizer. The values of $\beta_1$ and $\beta_2$ of the Adam optimizer are set to 0.9 and 0.999 respectively. The batch size is set to 16, and the low-resolution (LR) images have dimensions of 32$\times$32 pixels. To fine-tune the learning rate, we implement a cosine annealing learning strategy. Initially, the learning rate is set to $2\times10^{-4}$. The cosine annealing period for adjusting the learning rate spans 500,000 iterations. We have built all our models using the PyTorch framework and conducted the training on 4$\times$NVIDIA A800 GPUs. For our testing phase, we utilize several benchmark datasets, including Set5~\cite{Set5}, Set14~\cite{Set14}, BSD100~\cite{BSD100}, Manga109~\cite{Manga109}, Test2k \cite{kong2021classsr}, and Urban100~\cite{Urban100}. In addition, we also test our method on a realistic NTIRE 2018 SR challenge data \cite{timofte2017ntire} to further show our general applicability. For evaluation, we primarily evaluate the model's performance using the Peak Signal-to-Noise Ratio (PSNR), a commonly used metric for image quality assessment  \cite{jinjin2020pipal}. 

In our method, all the alignment operations are conducted before the last convolutional layer (i.e., the output layer) of the model. This setting holds true throughout all the experiments and baseline models used in this paper. We do this because we think aligning features at the end of the model propagation can most effectively regularize its behaviors to generate similar outputs for input images with the same content but different degradations. In addition, the Dropout ratios used for different baseline models in this paper follow the best setting of \citet{kong2022reflash} (i.e., SRResNet:0.7, RRDB:0.5, MSRN:0.5, SwinIR:0.5). More details of our implementation can be found in our codes. 

\section{Ablation Studies}
In this section, we show the ablation studies that verify the significance of our design. To be specific, we (1) review the design of brutly forcing the intermediate features of two images with identical contents but different degradations to be exactly the same, as discussed in \secref{sec:method}, and (2) justify the non-linear alignment design of our method. We run the experiments with SRResNet and RRDB on six benchmark datasets and use PSNR as the evaluation metric. The results are shown in \tableref{tab:ablation}. As we could observed, brutly forcing the features to be exactly the same, although theoretically the best, might put too much constraint on the model, limiting its ability to effectively reach a local minimum, thus yielding very unstable and unsatisfactory performances. On the other hand, experiments run with only linear alignment (i.e., w.o non-linear) show certain improvements, but its potential can be further excavated with the knowledge of higher dimension provided by the non-linear alignment.



\section{Detailed Comparisons with Dropout }
As we mentioned in \secref{sec:experiment}, we provide the detailed quantitative comparison results of \figref{fig:compare_drop2} in \tableref{table:degradations1}. Our method outperforms Dropout in almost all cases, which is not surprising and in line with our previous theoretical analyses.

\section{More Visual Results }
\begin{figure*}[b]
\centering
\includegraphics[width=1\linewidth]{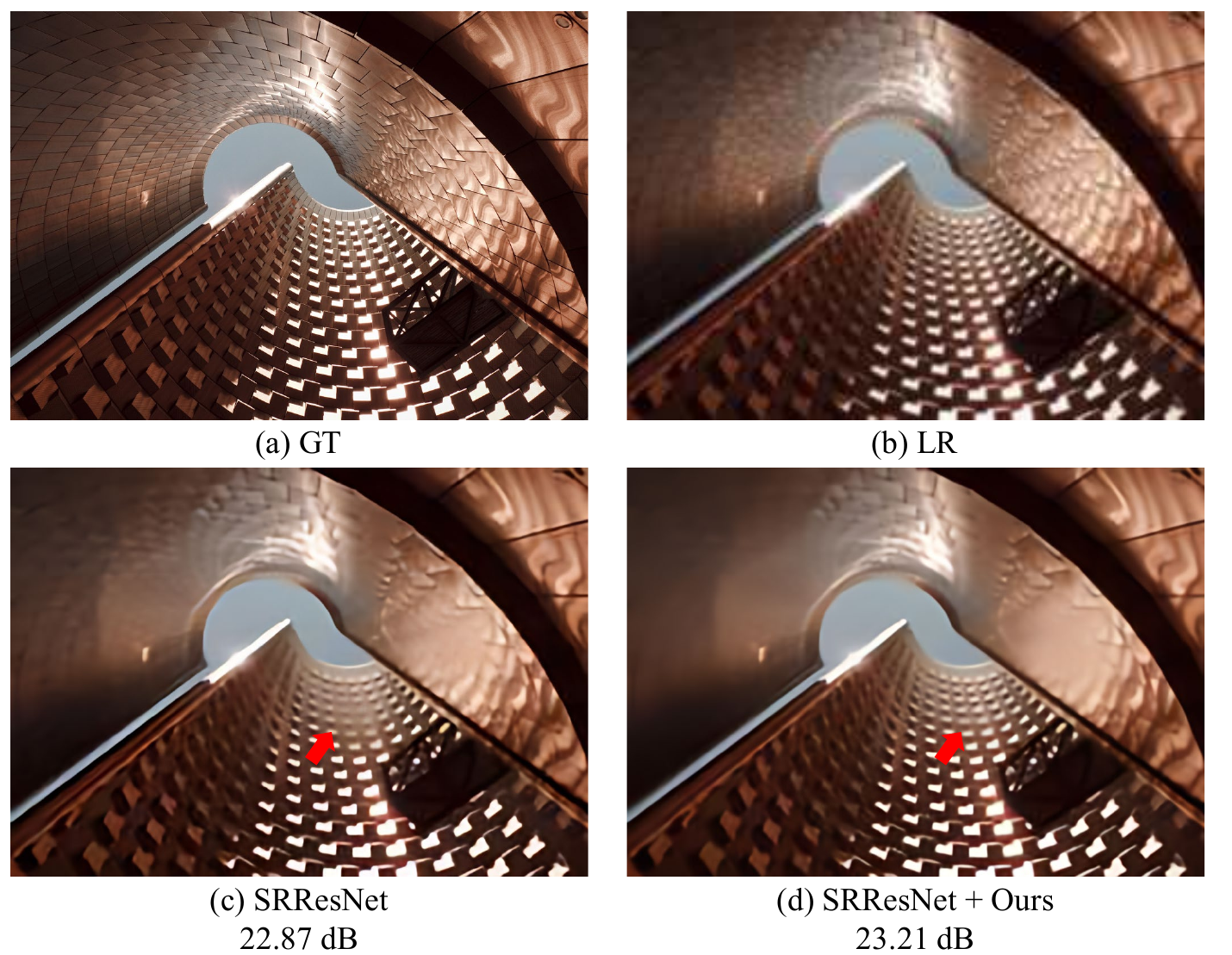}
\caption{\textbf{Visual comparison with and without our approach in “bicubic+noise20+jepg50”. (Zoom in for best view)
}}
\label{fig:compare1}
\end{figure*}

\begin{figure*}[h]
\centering
\includegraphics[width=1\linewidth]{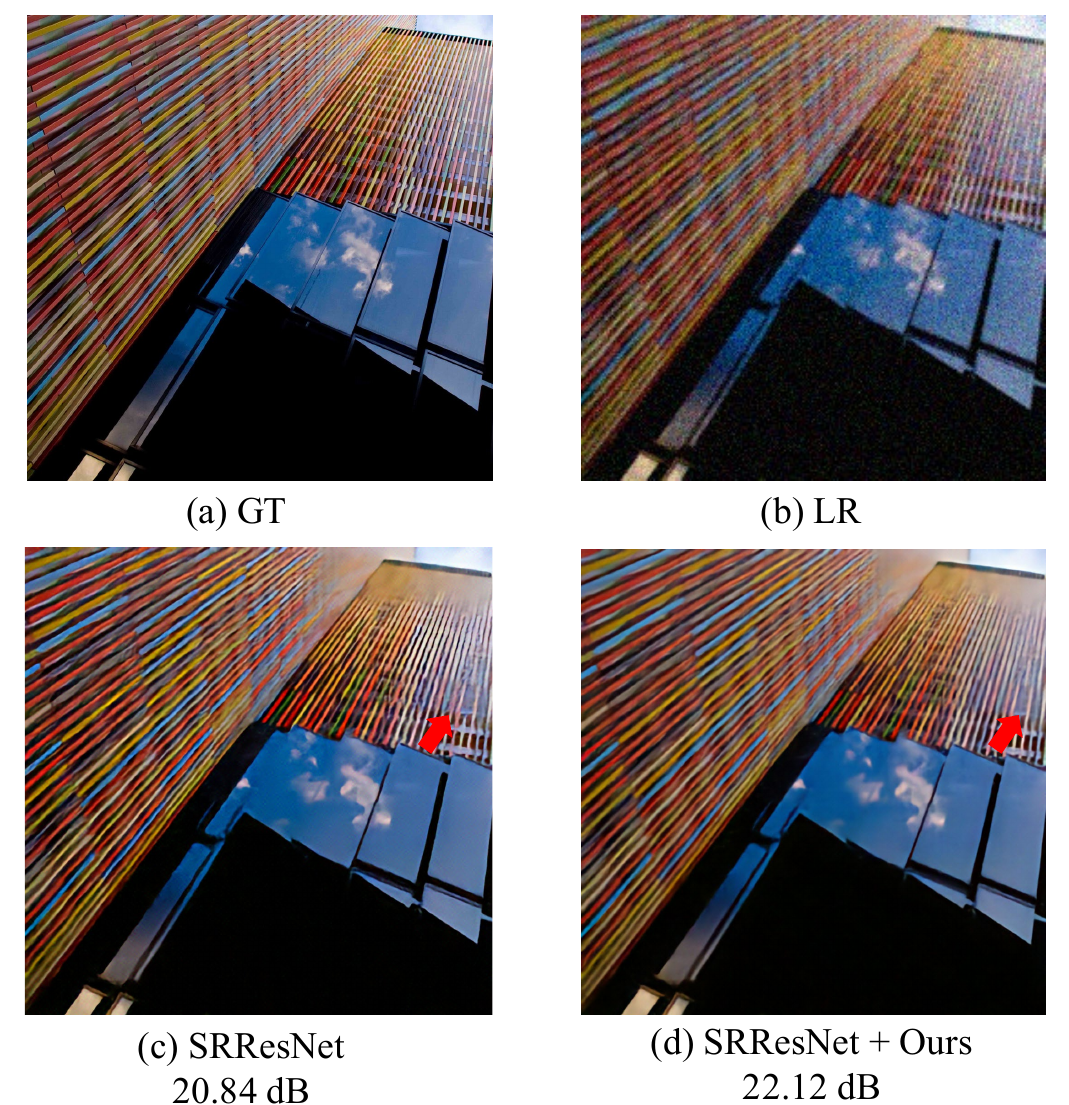}
\caption{\textbf{Visual comparison with and without our approach in “blur2+bicubic+jepg50”. (Zoom in for best view)
}}
\label{fig:compare2}
\end{figure*}

\begin{figure*}[h]
\centering
\includegraphics[width=1\linewidth]{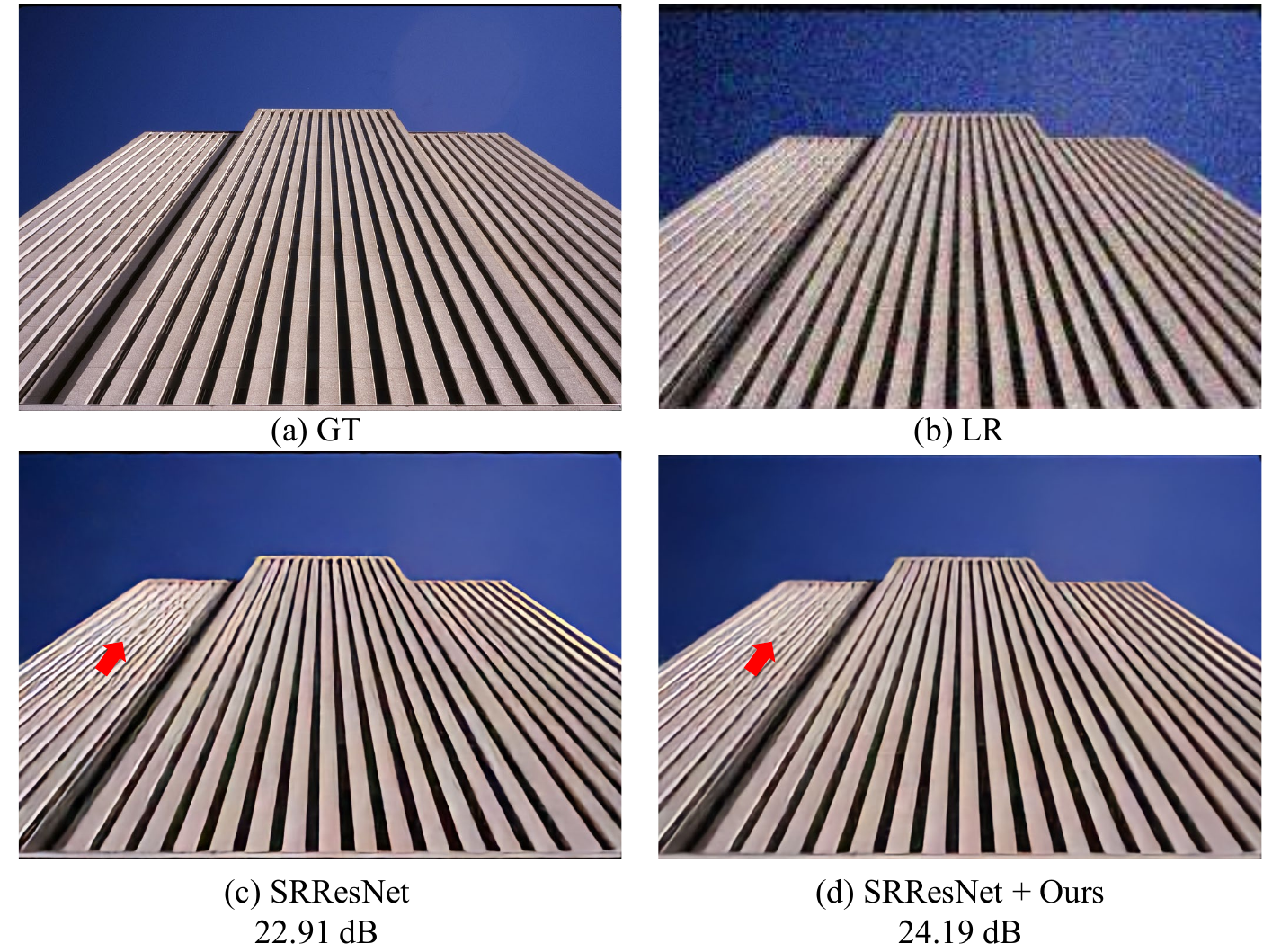}
\caption{\textbf{Visual comparison with and without our approach in “bicubic+noise20+jepg50”. (Zoom in for best view)
}}
\label{fig:compare3}
\end{figure*}

\begin{figure*}[h]
\centering
\includegraphics[width=1\linewidth]{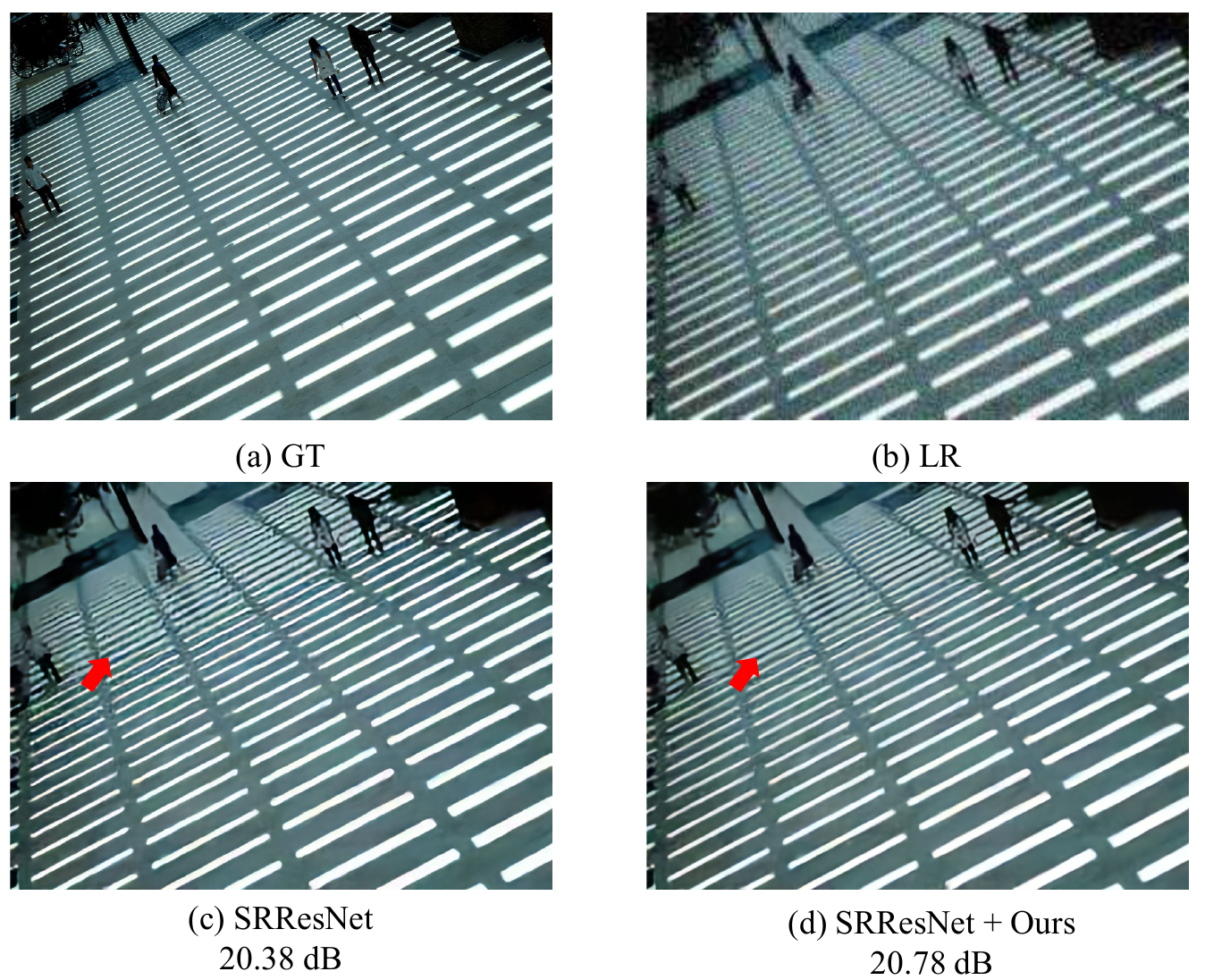}
\caption{\textbf{Visual comparison with and without our approach in “blur2+bicubic+noise20+jepg50”. (Zoom in for best view)
}}
\label{fig:compare4}
\end{figure*}

We provide more visual comparison results in \figref{fig:compare1}, \figref{fig:compare2}, \figref{fig:compare3}, and \figref{fig:compare4}. They are examples of different degradation restoration results (see the captions of the figures), and the red arrows in the figures highlight the main improvements of our method from human visual perspective.

\begin{table}[!t]
\setlength{\belowcaptionskip}{-1pt}
\renewcommand{\arraystretch}{1.1}
\caption{\textbf{Ablation Studies.}}\label{tab:ablation}
\centering
\setlength{\tabcolsep}{1.0mm}{
	\resizebox{1\linewidth}{!}{%
		\begin{tabular}{|l| l|}
			\hline
			\multirow{2}*{Models}  & \multicolumn{1}{c|}{PSNR $\uparrow$} \\
			\cline{2-2} & Set5 / Set14 / BSD  / Urban / Manga / Test2k \\
			\hline
			SRResNet & 23.53 / 22.23 / 22.34  / 20.49 / 18.40 / 22.95\\ 
			+brute-force & 23.49 / 22.28 / 21.94 / 20.27 / 18.97 / 22.93\\
			+w.o non-linear & 24.01 / 22.54 / 22.76 / 20.78 / 19.05 / 23.30\\
			+\textit{Ours} & \textcolor{red}{\textbf{24.20 / 22.83 / 22.82 / 20.96 / 19.12 / 23.41}} \\
			\hline
			RRDB &23.62 / 22.45 / 22.48  / 20.66 / 18.50 / 23.02\\
			+brute-force & 23.98 / 22.69 / 22.70 / 19.81 / 18.78 / 23.18\\
			+w.o non-linear & 24.44 / 22.94 / 23.45 / 20.97 / 19.02 / 23.39\\
			+\textit{Ours} &\textcolor{red}{\textbf{24.56 / 23.08 / 23.48 / 21.11 / 19.28 / 23.55}} \\
			\hline
		\end{tabular}%
	}
}
\end{table}

\begin{table*}[t]
\setlength{\abovecaptionskip}{-2pt}
\setlength{\belowcaptionskip}{-4pt}
\caption{\textbf{Six datasets with eight types of degradations (clean, noise, blur, jpeg, blur+noise, blur+jpeg, noise+jpeg, and blur+noise+jpeg) are used to evaluate the PSNR (dB) results of models with ×4 resolution.} The Dropout used in the experiments refers to the one in \citet{kong2022reflash}.}\label{table:degradations1}
\begin{center}
	\begin{tabular}{|l|cccc|cccc|cccc|}
		\hline
		\multirow{2}{*}{Models}                 & \multicolumn{4}{c|}{Set5~\cite{Set5}}                                                        & \multicolumn{4}{c|}{Set14~\cite{Set14}}                                                       & \multicolumn{4}{c|}{BSD100~\cite{BSD100}}                                                                                                      \\ \cline{2-13}
		&             \multicolumn{1}{c}{clean}                        & blur                         & \multicolumn{1}{c}{noise}                        & jpeg                         &  \multicolumn{1}{c}{clean}                        & blur                         & \multicolumn{1}{c}{noise}                        & jpeg                         & \multicolumn{1}{c}{clean}                     & blur            
		& \multicolumn{1}{c}{noise}                        & jpeg                 
		\\ \hline
		SRResNet \cite{SRResNet}         &       
		24.85&24.73&22.52&23.67&23.25&23.05&21.18&22.32&23.06&22.99&21.34&22.47 \\ 
		+ Dropout ($p$ = 0.7) & 25.63&25.23&22.79&24.05&23.73&23.45&21.23&22.62&23.31&23.26&21.30&22.69 \\
		+ Ours         & \bf 
		25.93& \bf25.62& \bf23.15& \bf24.38& \bf24.12& \bf23.80& \bf21.67& \bf22.99& \bf 23.83& \bf23.64& \bf21.77& \bf23.04\\ 
		\hline
		RRDB  \cite{ESRGAN}          &     
		25.18&25.12&21.79&23.82&23.74&23.36&21.02&22.59&23.38&23.32&21.00&22.73    \\ 
		+ Dropout ($p$ = 0.5) & 26.02&26.07&22.23&24.15&24.02&23.87&21.54&22.83&23.59&23.66&21.68&22.86\\
		+ Ours   & \bf 
		26.78& \bf26.55& \bf23.02& \bf24.70& \bf24.70& \bf24.35& \bf21.91& \bf23.21& \bf 24.59& \bf 24.54& \bf 23.47& \bf 23.67       \\ 
		\hline
		MSRN     \cite{li2018multi}           &   
		25.25&24.89&22.57&24.08&23.38&23.10&21.80&22.53&23.38&23.30&21.92&22.76\\ 
		+ Dropout ($p$ = 0.5) & 25.36&25.02&22.71&24.00&23.40&23.18&21.76&22.61&23.45&23.36&21.91&22.77\\
		+ Ours            & \bf 
		25.81& \bf25.52& \bf22.84& \bf24.46& \bf23.93& \bf23.64& \bf21.86& \bf22.83& \bf23.72& \bf23.58& \bf22.01& \bf22.98        \\ 
		\hline
		SwinIR     \cite{liang2021swinir}           &        26.25&26.03&22.96&24.37&24.53&24.25&22.08&23.14&23.91&23.83&22.12&23.04  \\ 
		+ Dropout ($p$ = 0.5)& 26.32&26.08&23.12&24.41&24.57&24.19&22.13&23.18&23.90&23.87&22.10&23.08\\
		+ Ours            & \bf 26.49& \bf26.23& \bf24.61& \bf24.68& \bf24.65& \bf24.28& \bf22.23& \bf23.29& \bf24.04& \bf23.96& \bf22.21& \bf23.15  \\ 
		
		\toprule\toprule
		& b+n                     & b+j                         & n+j                        & b+n+j                       & b+n                     & b+j                         & n+j                        & b+n+j    & b+n                     & b+j                         & n+j                        & b+n+j                      \\ \hline
		SRResNet \cite{SRResNet}         &   
		23.27&23.40&23.05&22.73&22.23&22.06&21.99&21.77&22.25&22.33&22.22&22.04     \\ 
		+ Dropout ($p$ = 0.7)
		&23.47&23.64&23.46&23.01&22.28&22.39&22.28&21.98&22.25&22.50&22.41&22.16 \\
		+ Ours         & \bf23.79& \bf23.86& \bf23.71& \bf23.19 & \bf22.65& \bf22.63& \bf22.55& \bf22.16& \bf22.53& \bf22.79& \bf22.62& \bf22.32\\ 
		\hline
		RRDB    \cite{ESRGAN}          &    
		23.44&23.45&23.32&22.81&22.47&22.17&22.29&21.95&22.39&22.47&22.42&22.15   \\ 
		+ Dropout ($p$ = 0.5) & 23.73&23.88&23.68&23.18&22.58&22.59&22.45&22.10&22.53&22.71&22.52&22.28 \\			
		+ Ours            & \bf 24.12& \bf24.14& \bf23.93& \bf23.26& \bf22.80& \bf22.76& \bf22.71& \bf22.21& \bf 22.85& \bf 23.21& \bf 22.97& \bf 22.54            \\ 
		\hline
		MSRN     \cite{li2018multi}           &   23.55&23.59&23.50&22.95&22.39&22.23&22.19&21.97&22.57&22.61&22.45&22.24    \\ 
		+ Dropout ($p$ = 0.5)& 23.73&23.61&23.52&23.04&22.43&22.26&22.24&21.96&22.59&22.64&22.44&22.20\\
		+ Ours            & \bf 23.70& \bf23.80& \bf23.73& \bf23.06& \bf22.52& \bf22.49& \bf22.48& \bf22.08& \bf22.68& \bf22.73& \bf22.56& \bf22.26           \\ 
		\hline
		SwinIR     \cite{liang2021swinir}           &23.80&23.84&23.67&22.99&22.53&22.73&22.59&22.20&22.61&22.82&22.61&22.34       \\
		+ Dropout ($p$ = 0.5) & 24.00&23.93&23.65&23.09&22.73&22.71&22.65&22.22&22.68&22.80&22.64&22.33 \\
		+ Ours            & \bf 24.13& \bf24.17& \bf23.89& \bf23.09& \bf22.87& \bf22.79& \bf22.81& \bf22.28& \bf22.77& \bf22.98& \bf22.76& \bf22.40        \\ 
		\bottomrule
		\toprule
		\multirow{2}{*}{Models}                 & \multicolumn{4}{c|}{Test2k~\cite{kong2021classsr}}                                                        & \multicolumn{4}{c|}{Urban100~\cite{Urban100}}                                                       & \multicolumn{4}{c|}{Manga109~\cite{Manga109}}                                                                                                      \\ \cline{2-13}
		&             \multicolumn{1}{c}{clean}                        & blur                         & \multicolumn{1}{c}{noise}                        & jpeg                         &  \multicolumn{1}{c}{clean}                        & blur                         & \multicolumn{1}{c}{noise}                        & jpeg                         & \multicolumn{1}{c}{clean}                     & blur            
		& \multicolumn{1}{c}{noise}                        & jpeg                 
		\\ \hline
		SRResNet \cite{SRResNet}         &     23.91&23.71&21.77&23.11&21.23&21.06&19.74&20.60&18.42&18.75&18.08&18.27 \\
		+ Dropout ($p$ = 0.7) & 24.26&23.98&21.75&23.27& 21.57&21.25&19.75&20.90&18.98&19.12&18.42&18.66  \\
		+ Ours         & \bf  24.58& \bf24.43& \bf22.17& \bf23.65& \bf21.94& \bf21.65& \bf20.19& \bf21.20 & \bf19.18& \bf19.46& \bf18.90& \bf19.02\\ 
		\hline
		RRDB    \cite{ESRGAN}          &     24.16&23.64&21.34&23.36 &21.57&21.18&19.61&20.93   & 18.59&18.64&18.30&18.41 \\ 
		+ Dropout ($p$ = 0.5) & 24.55&24.39&21.92&23.53 &21.89&21.75&19.92&21.12&18.73&19.03&18.72&18.60 \\
		+ Ours            & \bf 24.97& \bf24.76& \bf22.15& \bf23.86& \bf22.29& \bf21.95& \bf20.21& \bf21.40& \bf19.40& \bf19.61& \bf18.96& \bf19.24                 \\ 
		\hline
		MSRN     \cite{li2018multi}           &   22.99& 23.83 &22.30&23.22 & 21.35&21.14&20.19&20.75 &19.12&19.31&18.72&18.89\\ 
		+ Dropout ($p$ = 0.5) & 23.94&23.97&22.31&23.33&21.46&21.25&20.18&20.81&19.16&19.31&18.78&18.94 \\
		+ Ours            & \bf    24.52& \bf24.23& \bf22.38& \bf23.56& \bf21.88& \bf21.54& \bf20.22& \bf21.14& \bf19.23& \bf19.35& \bf18.84& \bf19.01     \\ 
		\hline
		SwinIR     \cite{liang2021swinir}           &        24.78 &24.57&22.71&23.63&22.18&21.90&20.56&21.32 &19.10&19.27&18.71&18.95 \\
		+ Dropout ($p$ = 0.5) & 24.81&24.54&22.76&23.69&22.27&21.99&20.67&21.38&19.15&19.30&18.73&19.03 \\ 
		+ Ours            & \bf  24.98& \bf24.76& \bf22.84& \bf23.80& \bf22.34& \bf22.07& \bf20.69& \bf21.48& \bf19.24& \bf19.45& \bf18.98& \bf19.28         \\ 
		\hline
		\toprule\toprule
		& b+n                     & b+j                         & n+j                        & b+n+j                       & b+n                     & b+j                         & n+j                        & b+n+j    & b+n                     & b+j                         & n+j                        & b+n+j                      \\ \hline
		SRResNet \cite{SRResNet}         &   22.81&22.87&22.85&22.59  &20.46&20.30&20.42&20.09&18.59&18.50&18.21&18.39 \\ 
		+ Dropout ($p$ = 0.7) & 22.89&23.11&23.04&22.70& 20.48&20.49&20.66&20.22&18.94&18.85&18.66&18.72 \\
		+ Ours         & \bf  23.11& \bf23.27& \bf23.22& \bf22.82& \bf20.73& \bf20.72& \bf20.91& \bf20.37& \bf19.27& \bf19.17& \bf18.98& \bf19.01\\ 
		\hline
		RRDB    \cite{ESRGAN}          &    22.93&22.87&23.12&22.73 &20.57&20.40&20.74&20.24 &18.83&18.43&18.38&18.41 \\ 
		+ Dropout ($p$ = 0.5) & 23.02&23.26&23.17&22.79&20.53&20.70&20.84&20.33&19.15&18.81&18.59&18.71 \\
		+ Ours            & \bf   23.14& \bf23.37& \bf23.34& \bf22.82& \bf20.76& \bf20.85& \bf21.03& \bf20.38& \bf19.43& \bf19.31& \bf19.12& \bf19.15      \\ 
		\hline
		MSRN     \cite{li2018multi}            &   23.03&23.01&22.94&22.66 &20.65&20.43&20.56&20.19 &19.16&19.02&18.80&18.88   \\ 
		+ Dropout ($p$ = 0.5) & 23.05&23.09&22.96&22.68&20.69&20.45&20.62&20.22&19.21&19.18&18.87&18.89\\
		+ Ours      & \bf 23.21& \bf23.21& \bf23.18& \bf22.78& \bf20.76& \bf20.64& \bf20.89& \bf20.26& \bf19.19& \bf19.18& \bf18.92& \bf18.93              \\ 
		\hline
		SwinIR     \cite{liang2021swinir}           &   23.15&23.27&23.21&22.81 &20.89&20.79&20.98&20.45  &19.07&19.02&18.79&18.80    \\
		+ Dropout ($p$ = 0.5) & 23.23&23.31&23.26&22.82&20.92&20.91&20.96&20.55&19.12&18.98&18.75&18.84\\ 
		+ Ours            & \bf     23.35& \bf23.47& \bf23.45& \bf22.93& \bf21.02& \bf20.98& \bf21.12& \bf20.53& \bf19.37& \bf19.35& \bf19.15& \bf19.12          \\ 
		\hline
	\end{tabular}
\end{center}
\vskip -0.3cm 
\label{table:degradations}
\vskip -0.30cm
\end{table*}


\end{document}

%% file: preamble.tex
\usepackage[dvipsnames]{xcolor}
